\documentclass[10pt]{article}

\usepackage{fontspec}
\usepackage{xeCJK}
\setCJKsansfont{FandolHei-Regular.otf}
\setCJKmonofont{FandolFang-Regular.otf}
\usepackage[a4paper,margin=1in]{geometry}
\usepackage{amsmath,amssymb,bm}
\usepackage{graphicx}
\usepackage{booktabs}
\usepackage{multirow}
\usepackage{xcolor}
\usepackage{url}
\usepackage[hidelinks]{hyperref}
\usepackage{caption}

\newcommand{\upcite}[1]{\cite{#1}}
\newcommand{\cnenfigcaption}[2]{\caption{#1}}
\newcommand{\cnentablecaption}[2]{\caption{#1}}
\title{Towards robust multimodal 3D object detection via visual foundation models}
\author{
Ziying Song\textsuperscript{1,$\dagger$},
Lin Liu\textsuperscript{1,$\dagger$},
Hongyu Pan\textsuperscript{2},
Shaoqing Xu\textsuperscript{3},\\
Lei Yang\textsuperscript{4},
Mingzhe Guo\textsuperscript{1}, and
Caiyan Jia\textsuperscript{1,*}\\[0.6em]
\small \textsuperscript{1}School of Computer Science and Technology, Beijing Jiaotong University, Beijing 100044, China\\
\small \textsuperscript{2}Horizon Robotics, Beijing 100000, China\\
\small \textsuperscript{3}University of Macau, Macao 999078, China\\
\small \textsuperscript{4}Nanyang Technological University, Singapore 349562, Singapore\\[0.4em]
\small \textsuperscript{$\dagger$}These authors contributed equally.\\
\small \textsuperscript{*}Corresponding author: \texttt{cyjia@bjtu.edu.cn}
}
\date{}

\hypersetup{
  unicode=true,
  pdftitle={Towards robust multimodal 3D object detection via visual foundation models},
  pdfauthor={Ziying Song, Lin Liu, Hongyu Pan, Shaoqing Xu, Lei Yang, Mingzhe Guo, Caiyan Jia}
}

\begin{document}

\maketitle

\begin{abstract}
Multimodal 3D object detection is fundamental to robust perception in autonomous driving because it integrates complementary information from LiDAR and camera sensors. However, existing methods often fail to maintain robustness under out-of-distribution (OOD) corruptions caused by sensor noise, adverse weather, and environmental changes. To address this problem, we propose \textbf{RoboDistill}, a robust and generalizable multimodal 3D object detection framework that leverages visual foundation models (VFMs), such as the \textbf{Segment Anything Model (SAM)}. First, we introduce \textbf{SAM-AD}, a domain-specific pretraining strategy that fine-tunes SAM on autonomous-driving imagery to extract feature representations with rich semantic information. Second, we design the \textbf{AD Feature Pyramid Network (AD-FPN)} to refine and upsample SAM features at multiple scales for seamless fusion with LiDAR features. Third, we develop the \textbf{Depth-Guided Wavelet Attention (DGWA)} module, which suppresses high-frequency sensor noise while preserving critical contextual information. Finally, we introduce \textbf{KD Fusion}, in which the pretrained SAM-AD serves as a teacher that distills high-quality visual knowledge into a lightweight point-cloud network, thereby improving robustness under noisy conditions. Extensive experiments across 27 challenging OOD corruption settings show that \textbf{RoboDistill} generally delivers stronger or competitive detection performance and robustness relative to representative state-of-the-art methods. This work bridges the gap between VFMs and 3D object detection and advances robust multimodal perception for real-world autonomous-driving applications.
\end{abstract}


\medskip
\noindent\textbf{Funding:} This work was supported by the National Natural Science Foundation of China under Grant No. 62536001 (Key Program) and Grant No. 62576026 (General Program).

\begin{figure}[!t]
\centering
\includegraphics[width=\textwidth]{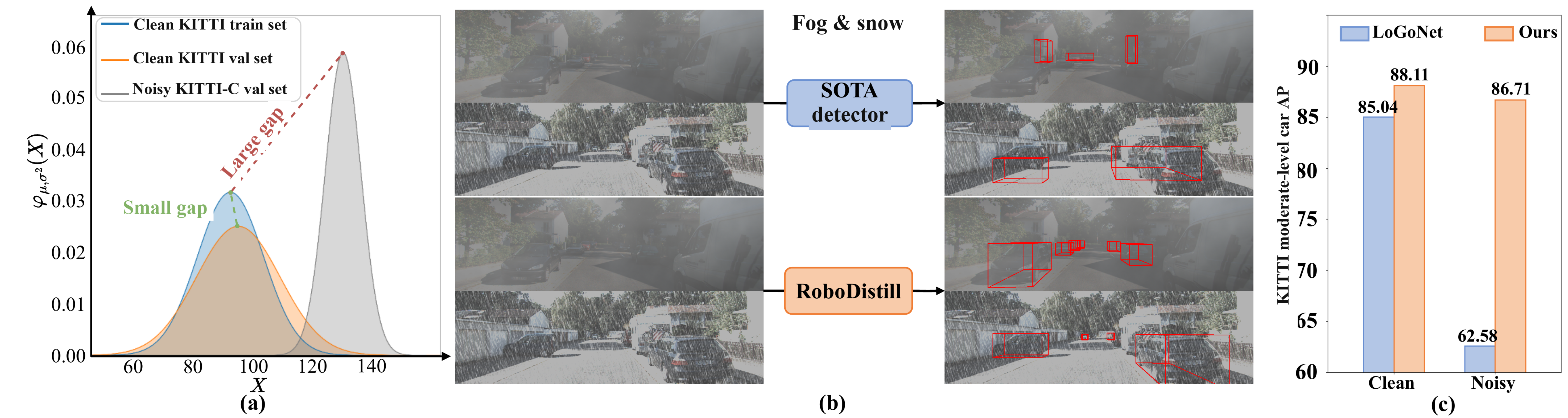}
\caption{\textbf{(a)} We use Gaussian distributions to characterize distributional differences across datasets. The results reveal a clear gap between the OOD-corrupted and clean validation sets. Specifically, the x-axis represents the set of mean pixel intensities in a dataset,
\(X=\{x_i\}_{i=1}^{N}\),
where
\(x_i=\frac{1}{H W 3}\sum_{h=1}^{H}\sum_{w=1}^{W}\sum_{c=1}^{3} I_{hwc}\).
Here, \(N\) is the number of samples, \(H\) and \(W\) denote the image height and width, and \(I_{hwc}\) is the pixel value.
\textbf{(b)} Visual foundation models, such as SAM \upcite{sam}, are robust to various corruptions; however, existing multimodal 3D detectors for autonomous driving remain vulnerable to OOD noise.
\textbf{(c)} RoboDistill integrates VFMs into a state-of-the-art multimodal 3D detection framework. Under the illustrated fog-and-snow corruption, RoboDistill achieves 86.71\% moderate-level car AP, compared with 62.58\% for LoGoNet~\upcite{logonet}, an improvement of 24.13 percentage points, while also performing better on the clean KITTI \upcite{kitti} dataset.
}
\label{fig:motivation}
\end{figure}

\section{Introduction}
In autonomous-driving scenarios, accurate and reliable perception is essential for tasks such as obstacle detection, object tracking, and trajectory planning \upcite{wang2023multi, oza2023unsupervised}. Unimodal systems based on either cameras or LiDAR often have limitations in complex environments. Camera-based systems provide rich semantic information but depend heavily on illumination and are vulnerable to adverse weather. In contrast, LiDAR-based systems provide precise geometric and depth information, but their measurements are sparse and semantically limited. By fusing the two modalities, multimodal methods can overcome the limitations of either modality and achieve robust perception. Multimodal 3D object detection has therefore become a cornerstone of robust perception systems because it integrates complementary information across modalities \upcite{wang2023multi, song2024robustness}.

Multimodal fusion has evolved substantially. Early methods such as MV3D \upcite{mv3d} demonstrated the feasibility of multimodal integration, followed by PointPainting \upcite{pointpainting} for semantic enhancement, TransFusion \upcite{transfusion} for cross-modal interaction, BEVFusion \upcite{bevfusion-mit} for unified representation, and CMT \upcite{cmt} for implicit spatial alignment. Although these state-of-the-art methods \upcite{song2023graphalign, robofusion, cmt, bevfusion-mit, GraphAlign_plus, xushaoqing_fusionpating, yin2024isfusion, song2024contrastalign} are robust in some respects, they are generally developed under idealized conditions and do not fully capture the complexity of real-world scenarios. For example, KITTI \upcite{kitti} primarily contains clear-weather scenes, lacks diverse conditions such as snow, fog, and heavy rain, and does not account for sensor noise, which is critical in practice. KITTI-C \upcite{Robustness3d} was therefore introduced to simulate various OOD corruption conditions. As shown in Figure~\ref{fig:motivation}, the OOD-corrupted and clean datasets exhibit a pronounced distribution gap. Consequently, state-of-the-art methods trained on clean data may overfit specific scenarios and fail to generalize to OOD environments such as nighttime driving, extreme weather, or severe sensor noise. This raises a central research question: \textbf{How can a method remain robust and reliable under unseen, challenging conditions?}

The gap between OOD scenarios and clean training data can be reduced by increasing dataset diversity to cover more real-world environments or by designing algorithms that are robust to sensor noise and environmental variation. However, constructing large-scale datasets spanning diverse conditions is difficult and expensive. A natural alternative is domain adaptation (DA), which is commonly used to bridge source and target domains \upcite{wang2023ssda3d, tsai2023viewer, ST3D, SPG}. Although DA can improve the robustness of 3D object detection by reducing dependence on large annotated datasets and adapting models to new domains, it has inherent limitations, including difficulty handling large domain gaps, label-distribution shifts, and the risk of overfitting \upcite{oza2023unsupervised}. When the source and target domains differ substantially, DA often generalizes poorly and degrades performance in the target domain.

Recent years have witnessed transformative advances in natural language processing and computer vision with the emergence of foundation models \upcite{sam, gpt4, fastsam, mobilesam, ma2024segment, Sam_adapter}. By serving as feature extractors, predictors, or interpreters, these models have established new paradigms in deep learning \upcite{moor2023foundation, fei2022towards}. Visual foundation models (VFMs), in particular, exhibit strong generalization owing to pretraining on large and diverse datasets \upcite{sam, fastsam, mobilesam}.
We emphasize that our use of SAM does not rely on its segmentation outputs themselves, but on its learned ``object-region consistency'' and ``boundary-structure priors''. These priors are generally more stable under rain- or fog-induced blur, partial occlusion, and background perturbations, providing high-signal-to-noise semantic anchors that (1) mitigate semantic drift caused by image-side noise and (2) facilitate 2D--3D alignment while reducing cross-modal mutual-information loss, thereby improving OOD robustness.
These advances motivate a new approach that uses VFMs to improve the robustness of multimodal 3D object detection. VFMs can serve as high-level semantic priors to facilitate cross-modal alignment and alleviate spatial misalignment and feature redundancy \upcite{seal}. We therefore leverage these models to address the challenges faced by multimodal 3D object detectors in OOD-corrupted scenarios.

Accordingly, we present a robust and generalizable framework---\textbf{RoboDistill}---that introduces VFMs into the multimodal 3D detection pipeline, enabling existing detectors to transfer effectively from clean, controlled settings to real-world autonomous-driving scenarios with complex corruptions.
RoboDistill forms a systematic pipeline of ``semantic anchoring $\rightarrow$ geometric denoising $\rightarrow$ controlled transfer'': SAM-AD provides stable semantic priors, AD-FPN aligns them to fusion-ready scales, DGWA suppresses noise propagation on the 3D side, and KD Fusion transfers reliable semantic knowledge to point-cloud representations to improve robustness.
First, because generic SAM is vulnerable to viewpoint changes, dynamic objects, and complex backgrounds in autonomous-driving scenes, we propose an autonomous-driving-specific domain-adaptive pretraining strategy that yields \textbf{SAM-AD}. It aligns VFM capabilities with the requirements of 3D detection and improves cross-modal feature alignment.
Second, because the resolution and hierarchy of VFM features make them difficult to integrate directly into a 3D detection pipeline, we design \textbf{AD-FPN} to efficiently upsample and refine SAM features. This module seamlessly fuses high-level semantics with LiDAR features while preserving both spatial resolution and semantic expressiveness.
Furthermore, to suppress corruption in OOD scenarios, we propose a \textbf{Depth-Guided Wavelet Attention module (DGWA)}. It decomposes LiDAR depth features into high-frequency noise and low-frequency contextual components and uses attention to selectively suppress noise while retaining critical structural information, yielding more stable representations under fog, rain, and low illumination.
Compared with spatial-domain filtering or normalization, frequency-domain decomposition disentangles ``nonstationary noise'' from ``structured geometry'' across multiple scales. Correlation analysis between high-frequency coefficients and noise intensity, frequency-band ablations (low-frequency only/high-frequency only/full spectrum), and comparisons with spatial-domain denoising baselines jointly support this design.
Finally, we propose \textbf{KD Fusion}, a multimodal knowledge-distillation module based on a teacher--student paradigm. \textbf{SAM-AD} serves as the teacher, providing high-quality image representations that are distilled into a lightweight point-cloud student network. The student thereby inherits large-scale pretrained knowledge and domain-generalization capability, improving feature extraction and fusion under corruption and OOD conditions.
To account for the 2D--3D domain gap, the distillation loss transfers guidance from the teacher to the student only during training. We distill \emph{task-aligned predictive distributions} to avoid bias from rigid feature-space alignment and introduce teacher-confidence gating (high entropy$\rightarrow$weak distillation) to suppress unreliable soft labels in corrupted scenarios, thereby reducing negative transfer while retaining stable gains.
In summary, pretraining on large and diverse datasets gives VFMs strong generalization, efficient feature extraction, and robust semantic understanding even in challenging environments. By integrating VFMs into the multimodal 3D detection pipeline, RoboDistill addresses cross-modal alignment, noise suppression, and feature fusion, providing a robust and scalable solution for autonomous-driving applications.

\section{Related Work}\label{sec2}

\subsection{Multimodal 3D Object Detection}
Multimodal 3D object detection has attracted considerable attention because it exploits complementary information from different sensors, particularly on popular datasets such as KITTI \upcite{kitti} and nuScenes \upcite{nuscenes}. Images provide rich semantic cues such as color and texture, whereas LiDAR point clouds capture precise depth and geometry. Existing studies \upcite{autoalignv2, deepinteraction, focalconv, logonet, sparsefusion, song2023graphalign, GraphAlign_plus} emphasize deep cross-modal feature integration. For example, PointPainting \upcite{pointpainting} enriches LiDAR point-cloud representations with semantic features extracted by a pretrained 2D segmentation network. Similarly, PointAugmenting \upcite{pointaugmenting} and AutoAlignV2 \upcite{autoalignv2} propose more efficient cross-modal interaction strategies that use global image features to improve point-cloud processing. Although these methods improve accuracy, robustness to real-world corruption and environmental variation remains challenging. In recent years, bird's-eye-view (BEV) representations have become the dominant paradigm for multimodal 3D object detection. BEV-based methods \upcite{bevfusion-mit, cmt, song2025graphbev, chen2023futr3d, yin2024isfusion}, such as BEVFusion \upcite{bevfusion-mit}, unify multimodal features in BEV space for efficient fusion and improved spatial reasoning. Notably, GraphBEV \upcite{song2025graphbev} introduces graph-structured, depth-aware feature alignment to address inaccurate point-cloud projection.
Although existing state-of-the-art multimodal methods perform well on clean, controlled datasets, they often fail to account for real-world complexities such as adverse weather, sensor noise, and domain shifts \upcite{song2024robustness, Robustness3d}. For example, datasets such as KITTI and nuScenes lack diversity in weather and environmental conditions, leading to overfitting and reduced generalization in OOD scenarios.

\subsection{Applications of Visual Foundation Models in Computer Vision}
Alongside the rapid development of large language models \upcite{Flan-T5}, numerous visual foundation models (VFMs) have emerged in computer vision \upcite{sam, fastsam, mobilesam, XDecoder}. Through large-scale pretraining on diverse datasets, these VFMs generalize well to new visual scenarios and extract pixel-level features. However, most VFMs focus on 2D vision, and their extension to 3D perception remains underexplored, leaving substantial room to adapt or extend existing 2D VFMs to 3D tasks. The first VFM, SAM \upcite{sam}, is built on the Vision Transformer (ViT) \upcite{vit} and trained on the SA-1B dataset containing 11 million samples, yielding strong scene-level generalization. FastSAM \upcite{fastsam} provides a real-time CNN solution that greatly reduces computational cost while retaining performance. MobileSAM \upcite{mobilesam} distills SAM's large image encoder (ViT-H) into a lightweight encoder compatible with SAM's mask decoder. As general-purpose large models, these VFMs provide powerful tools for downstream applications. Despite progress in 2D vision, VFM research in 3D remains nascent. SAM3D \upcite{SAM3D}, a LiDAR-only method, projects 3D LiDAR data into 2D BEV space to exploit SAM's generalization, but its limited modal interaction constrains performance. Compared with the previous RoboFusion method, RoboDistill uses SAM for multimodal knowledge distillation to transfer and learn model knowledge.
In summary, existing state-of-the-art multimodal 3D object detectors remain challenged by OOD-corrupted scenarios and struggle to bridge the gap between ``clean'' training data and OOD corruption. The generalization and robustness of VFMs create new opportunities for visual tasks, motivating us to use these capabilities to address OOD-corruption generalization in multimodal 3D detection.

\section{RoboDistill}\label{sec3}
\begin{figure}[!t]
\centering
\includegraphics[width=\textwidth]{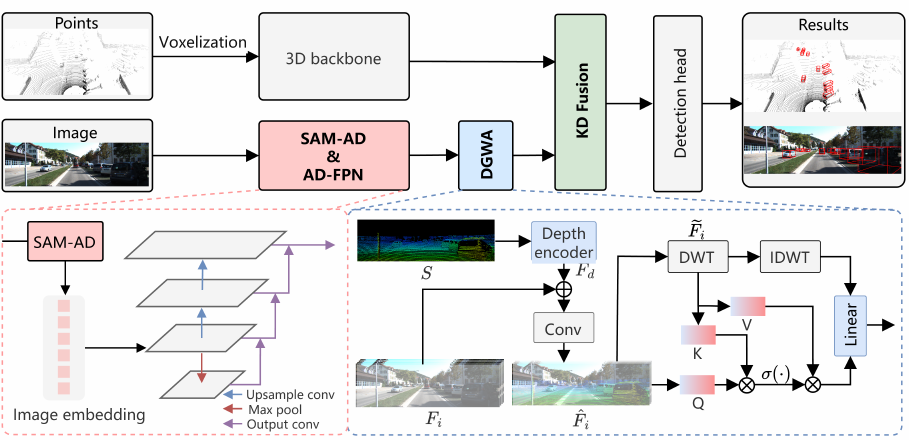}
\caption{Overall framework of RoboDistill.}
\label{fig:framework}
\end{figure}

In this section, we present the \textbf{RoboDistill} framework, illustrated in Figure~\ref{fig:framework}. Extending our preliminary work, RoboDistill introduces a new multimodal strategy that fully exploits the generalization capability of visual foundation models (VFMs). Unlike prior methods, \textbf{RoboDistill} combines VFMs with robust cross-modal alignment to explicitly address out-of-distribution (OOD) corruption in multimodal 3D object detection, offering a new solution for robust and scalable perception in real-world autonomous-driving scenarios. Our detection head follows that of Voxel R-CNN \upcite{voxelrcnn}.

\subsection{SAM-AD \& AD-FPN}
As a representative visual foundation model, SAM~\upcite{sam} exhibits strong generalization through pretraining on the large-scale SA-1B dataset. The dataset contains more than 11 million samples and 1 billion high-quality masks, giving SAM strong robustness and adaptability across diverse visual scenarios. Current SAM-family models~\upcite{sam, fastsam, mobilesam} primarily support 2D vision tasks. Directly extending VFMs such as SAM to 3D tasks is challenging because 2D and 3D tasks differ fundamentally in data representation and task requirements. To bridge this gap, we combine SAM with a multimodal 3D model, fusing robust 2D representations with 3D point-cloud features to obtain more robust fused representations.

\begin{figure}[!t]
\centering
\includegraphics[width=\textwidth]{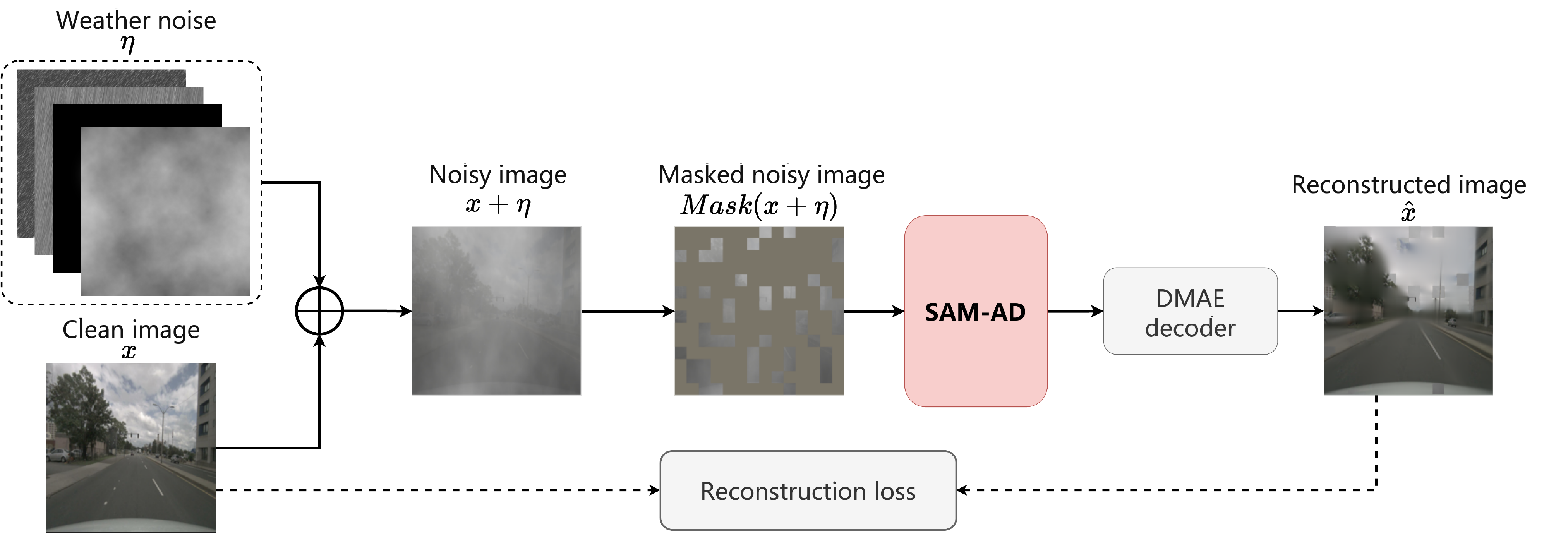}
\caption{Illustration of the pretraining framework. We corrupt a clean image $x$ with a noise perturbation $\eta$ to obtain $x+\eta$. We then randomly mask several patches in the corrupted image to obtain $Mask(x+\eta)$. The SAM-AD encoder and DMAE decoder are trained to reconstruct the corresponding clean image $\hat{x}$ from $Mask(x+\eta)$.}
\label{fig:pretrain}
\end{figure}

\noindent \textbf{SAM-AD Module. }
To better adapt SAM to autonomous-driving scenarios, we perform domain-adaptive pretraining to obtain SAM-AD, which is optimized for autonomous-driving tasks. Specifically, we collect a large number of autonomous-driving images from established datasets such as KITTI~\upcite{kitti} and nuScenes~\upcite{nuscenes}, covering diverse driving scenes and environmental conditions, to construct a base AD image dataset. During pretraining, we follow the DMAE framework~\upcite{dmae} and use masked image modeling to pretrain SAM and MobileSAM in a self-supervised manner. Specifically, as shown in Figure~\ref{fig:pretrain}, let $x$ be a clean image sampled from the AD dataset and let $\eta$ denote the set of corrupted images obtained by applying the perturbations in~\upcite{Robustness3d} to $x$. Because autonomous-driving systems often encounter OOD disturbances such as complex weather, illumination changes, and imaging degradation in real deployments, we jointly model all corruption types defined in Robustness3D\upcite{Robustness3d} during pretraining. For example, the nuScenes setting contains 27 distinct corruption types, each with severity levels 1--5, spanning distribution shifts from mild to severe degradation. Instead of training a separate model for each corruption type, we adopt unified mixed-corruption pretraining for a shared large model, thereby learning more general and robust visual representations. For FastSAM, we pretrain its segmentation head on the AD dataset using YOLOv8. To prevent overfitting, we apply random scaling and cropping, with a mask ratio of 0.75. All models are trained on 8 NVIDIA A100 GPUs for 400 epochs. We evaluate reconstruction accuracy and downstream 3D object-detection metrics; the results demonstrate that SAM-AD is robust and generalizes well in OOD scenarios.

\noindent \textbf{AD-FPN Module. }
As a promptable segmentation model, SAM consists of an image encoder, a prompt encoder, and a mask decoder. The prompt encoder and mask decoder primarily serve semantic segmentation, whereas we use the image encoder to extract robust, high-quality image embeddings. SAM employs a ViT (Vision Transformer)~\upcite{vit}-based image encoder that produces high-dimensional, low-resolution embeddings with a stride of 16 (i.e., scale=$1/16$). These features, however, lack multiscale representation, hindering accurate localization and detection when objects vary in size.

To address this issue, we design AD-FPN based on the FPN~\upcite{fpn} principle to enhance the multiscale expressiveness of ViT embeddings. Specifically, AD-FPN takes a ViT embedding as input and constructs multiscale features $F_s$ at strides $s \in \{32,16,8,4\}$. Each feature map $F_s \in \mathbb{R}^{\frac{H}{s} \times \frac{W}{s} \times C_s}$ is progressively refined in a bottom-up manner. Unlike a conventional FPN, AD-FPN is tailored to ViT embeddings, improving spatial resolution and fine-grained representation while retaining semantic robustness. This enhancement enables more accurate detection of small, distant, or occluded objects and substantially alleviates common visual challenges in autonomous-driving scenes. AD-FPN also improves the robustness and adaptability of the multimodal perception pipeline under adverse conditions, such as low illumination or dense traffic, providing reliable and efficient features for downstream 3D detection and perception.

\subsection{Depth-Guided Wavelet Attention Module (DGWA)}

Although SAM-AD and SAM can extract robust image features, a substantial gap remains between the 2D and 3D feature domains. In corrupted environments, images often amplify noise because cameras lack geometric priors, leading to negative feature transfer. We therefore propose the \textbf{Depth-Guided Wavelet Attention module (DGWA)}, which improves feature robustness through two key operations: ``depth-guided feature enhancement'' and ``wavelet-domain noise suppression''.

\noindent \textbf{Depth-Guided Feature Enhancement. }
This module introduces geometric priors into image features by fusing image and depth features. At the stride-4 fusion level, let the image feature be \(F_i \in \mathbb{R}^{\frac{H}{4} \times \frac{W}{4} \times C_i}\) and the depth feature be \(F_d \in \mathbb{R}^{\frac{H}{4} \times \frac{W}{4} \times C_i}\). Here, $F_d$ is obtained by projecting the point cloud into the image coordinate system and processing the resulting sparse LiDAR depth map \(S \in \mathbb{R}^{H \times W \times 2}\) with a depth encoder. A convolutional fusion operation then produces the depth-guided feature $\hat{F_i}=Conv(Concat\{F_i, F_d\})$, embedding spatial and geometric information into the representation and substantially improving its resistance to noise.

\noindent \textbf{Wavelet-Domain Noise Suppression. } We use the Haar wavelet as the default basis for DGWA because its decomposition is simple, computationally efficient, and effective at separating low-frequency structural information from high-frequency noise, making it well suited to real-time feature enhancement for robust autonomous-driving perception.
For the enhanced feature \(\hat{F_i}\), a Haar wavelet transform decomposes it into four subbands: one low-frequency subband \(\widetilde{f}_i^{LL} \in \mathbb{R}^{\frac{H}{8} \times \frac{W}{8} \times C_i}\) and three high-frequency subbands \((\widetilde{f}_i^{LH}, \widetilde{f}_i^{HL}, \widetilde{f}_i^{HH}) \in \mathbb{R}^{\frac{H}{8} \times \frac{W}{8} \times C_i}\). The low-frequency subband retains coarse contextual information, whereas the high-frequency subbands capture fine-grained edges and textures, facilitating identification of noise signals. Concatenating these subbands yields the wavelet feature \(\widetilde{F_i} \in \mathbb{R}^{\frac{H}{8} \times \frac{W}{8} \times 4C_i}\), which is weighted by the wavelet-attention mechanism \(Att_{\omega}\). This mechanism selectively suppresses noise while preserving useful information:
\begin{align}
F_{att} = Att_{\omega}(\hat{F_i}, \widetilde{F_i}) = \sigma\left(\frac{\hat{F_i}W^q (\widetilde{F_i}W^k)^T}{\sqrt{C_i}}\right) \widetilde{F_i}W^v.
\end{align}

Finally, the inverse discrete wavelet transform (IDWT) reconstructs the features, which are fused with the attention output to produce the final robust representation \(F_{out}\):
\begin{align}
   F_{out} = MLP(\text{Concat}(F_{att}, \hat{F_i})).
\end{align}

This design enables \(F_{out}\) to suppress redundant noise while retaining critical information in both the spatial and frequency domains, thereby improving robustness and generalization in downstream tasks such as object detection and cross-modal feature fusion.

\subsection{Multimodal Knowledge-Distillation Fusion (KD Fusion)}
\label{sec:fusion}

\begin{figure}[!t]
\centering
\includegraphics[width=0.7\textwidth]{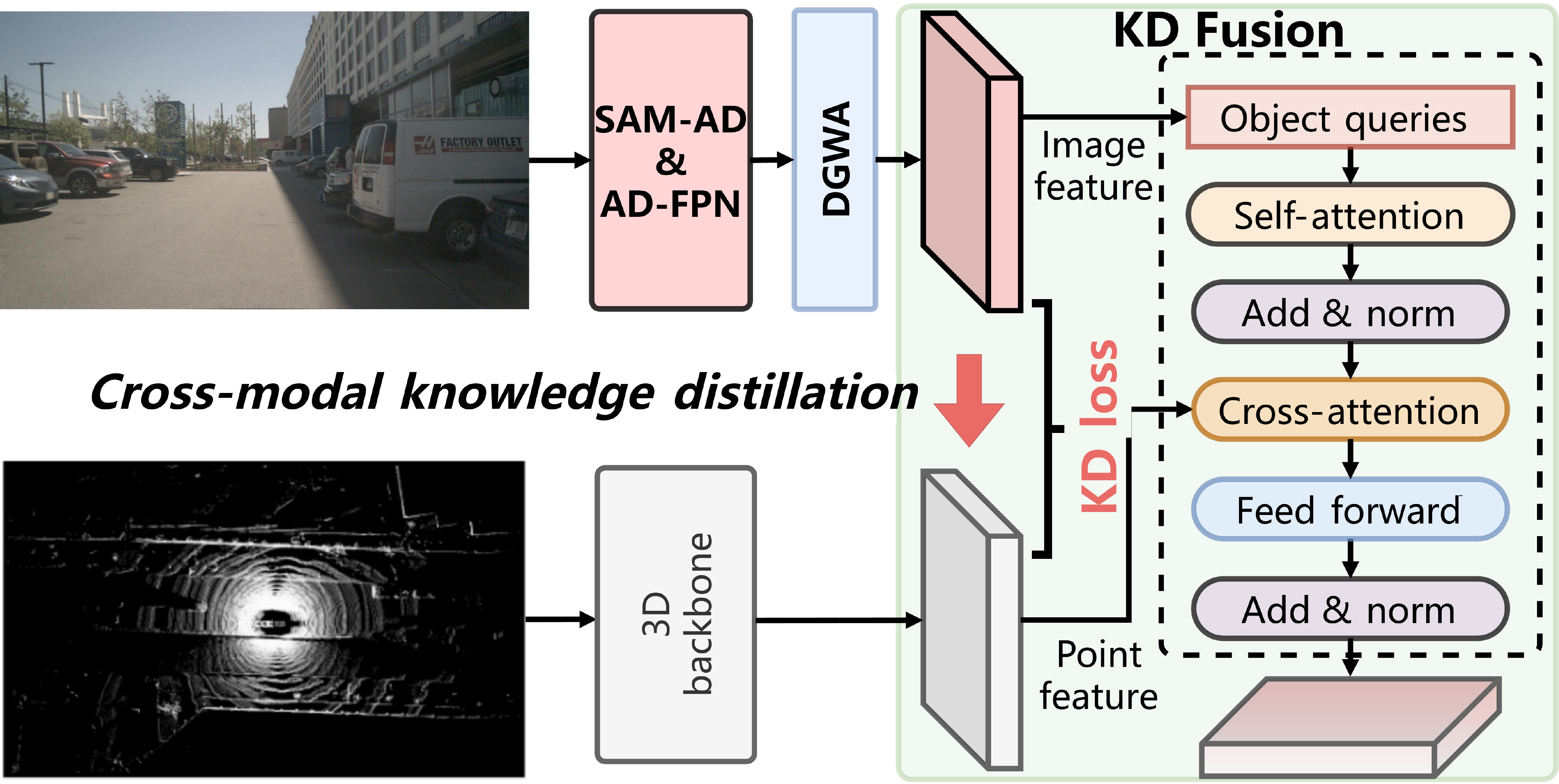}
\caption{Architecture of \textbf{KD Fusion}. The image branch uses SAM-AD as the teacher model, whereas the point cloud branch serves as the student model. \textbf{KD Fusion} transfers knowledge from the teacher to the student and uses a Transformer architecture for multimodal feature alignment and fusion.}
\label{fig:kdfusion}
\end{figure}

Although SAM-AD and SAM can extract robust image features, bridging the 2D--3D domain gap in multimodal fusion remains a major challenge. Under data heterogeneity, directly aligning 2D and 3D modalities often fails to preserve mutual information, resulting in suboptimal representations. Moreover, SAM-processed image features are more robust and transferable because of SAM's strong generalization. We therefore designate the image branch as the \textbf{teacher model} and the point-cloud branch as the \textbf{student model}, enabling cross-modal knowledge transfer that improves the student's learning capability. As shown in Figure~\ref{fig:kdfusion}, the proposed \textbf{multimodal knowledge-distillation fusion module (KD Fusion)} exploits the complementarity of the teacher--student structure to improve cross-domain robustness.
During distillation, the teacher provides soft targets and robust priors; during multimodal inference, the image and point-cloud branches jointly participate in feature fusion.
The module comprises two core components: cross-modal knowledge transfer and Transformer fusion.

\noindent \textbf{Cross-Modal Knowledge Transfer. }
The SAM-AD teacher extracts rich semantic information from the input image \(I \in \mathbb{R}^{H \times W \times 3}\) and uses the refined image feature $F_{out}$ to generate classification logits \(T \in \mathbb{R}^{N \times C}\). Meanwhile, the student extracts features from the point-cloud data \(P \in \mathbb{R}^{N \times D}\) and generates logits \(S \in \mathbb{R}^{N \times C}\).
For optimization, we first complete domain-adaptive pretraining of SAM-AD and then optimize a joint objective of ``\textbf{supervised learning (CE) + distillation learning (KL)}'': the CE term maintains a lower bound on task accuracy, whereas the KL term provides cross-modal class relationships and robust priors, balancing lightweight design with performance retention.

To reduce the modal gap while preserving task accuracy, we design a knowledge-distillation loss that encourages the student's predictive distribution to approach the teacher's. Given the substantial domain gap between 2D semantic space and 3D point-cloud space, we distill in the \textbf{task-aligned logit/probability space}, rather than directly enforcing intermediate-feature alignment, to reduce the risk of negative transfer caused by cross-domain mismatch.
The loss is a weighted combination of the Kullback--Leibler (KL) divergence between the teacher and student predictive distributions and the cross-entropy loss of the predictions:
\begin{align}
\mathcal{L}_{KD}
=
\alpha \cdot \underbrace{w(T)\,\tau^{2}\,\mathrm{KL}\!\left(
\mathrm{softmax}\!\left(\tfrac{T}{\tau}\right)\, \big\| \,\mathrm{softmax}\!\left(\tfrac{S}{\tau}\right)
\right)}_{\text{distill from teacher}}
+ (1 - \alpha) \cdot \mathcal{L}_{CE}(S, Y),
\end{align}
Here, $\tau$ is the temperature coefficient that smooths the distillation distribution to transfer interclass similarities, and $\alpha$ balances the distillation and supervised terms. We fix $\alpha=0.5$ and $\tau=4$ in all experiments to improve training stability and reproducibility.

To avoid negative transfer under severe corruption or OOD conditions, we introduce a teacher-confidence weight $w(T)$ that adaptively controls distillation strength. When the teacher prediction is unreliable (high entropy/low confidence), distillation is weakened or disabled to prevent erroneous soft labels from misleading the student. A simple implementation uses normalized-entropy gating:
$w(T)=\max\!\left(0, 1-\tfrac{\mathcal{H}(\mathrm{softmax}(T))}{\log C}\right)$, where $\mathcal{H}(\cdot)$ denotes information entropy.

With KD Fusion, the student learns more robust and compact 3D representations under the guidance of the SAM-AD teacher, enabling it to handle noise and uncertainty in point-cloud data more effectively.

Qualitatively, SAM-AD provides guidance through two main components, ``semantic priors + interclass structure'': (i) semantic priors derived from large-scale pretraining and AD-domain adaptation help maintain stable class-discriminative cues under fog, rain, or low illumination; and (ii) interclass structure is transferred through a temperature-smoothed soft distribution, constraining the student against overconfidence and decision-boundary drift under corruption. Together with confidence gating, the teacher imposes constraints only within its ``reliable regions'', effectively reducing the risk of negative transfer when domain gaps and corruption coexist.

\noindent \textbf{Transformer-based fusion. }
To efficiently fuse image and point-cloud features, KD Fusion employs a Transformer-based fusion operator that combines cross-attention with learnable object queries. This design enables the point-cloud network (student) to inherit robust representations from the image network (teacher) through cross-modal knowledge distillation. The fusion operator comprises three key stages: task-specific self-attention refinement, cross-modal alignment via cross-attention, and residual feed-forward refinement.

\begin{itemize}
    \item
    \textbf{Self-attention for task-specific refinement. }
    The fusion module uses learnable object queries to guide feature refinement. These queries serve as embeddings associated with the target task, such as object detection. First, self-attention models dependencies among the queries and strengthens their focus on task-relevant features:
    \begin{align}
    \text{Attention}(Q, K, V) = \text{Softmax}\left(\frac{QK^T}{\sqrt{d}}\right)V,
    \end{align}
    where \(Q\), \(K\), and \(V\) are the query, key, and value matrices, respectively, generated from the image features $F_{out}$. This process allows the object queries to adaptively learn task-specific representations.

    \item
    \textbf{Cross-attention for multimodal alignment. }
    After self-attention refinement, the object queries further interact with the image and point-cloud features. Cross-attention aligns semantically rich image features with spatially detailed point-cloud features, thereby integrating complementary information across modalities:
    \begin{align}
    \text{CrossAttention}(Q, K, V) = \text{Softmax}\left(\frac{QK^T}{\sqrt{d}}\right)V,
    \end{align}
    where \(Q\) denotes the query vectors derived from the image features $F_{out}$, while \(K\) and \(V\) are the key and value matrices, respectively, generated from the point-cloud features. Through this mechanism, the student model (point-cloud network) can effectively inherit semantic knowledge from the teacher model (image network).

    \item
    \textbf{Residual feed-forward refinement. }
    To stabilize the fusion process, residual connections and layer normalization are applied in every layer. A feed-forward network (FFN) then further improves the nonlinear expressiveness of the fused features, thereby enhancing robustness in downstream tasks.
\end{itemize}

The Transformer-based fusion operator jointly integrates semantic, spatial, and geometric information from the image and point-cloud modalities. By aligning heterogeneous modalities through cross-attention, it facilitates efficient multimodal feature interaction and substantially improves robustness to noisy or incomplete inputs through knowledge distillation.
\begin{table}[t]
\centering
\caption{Comparison of RoboDistill (L/B/T) with representative state-of-the-art methods, including recent methods published since 2025, on the \textbf{KITTI} validation and test sets in terms of car-class AP$_{3D}$ (R$_{40}$).}
\renewcommand\arraystretch{0.80}
\tabcolsep=5.99mm
\resizebox{\linewidth}{!}
{
\begin{tabular}{l|cccc|cccc}
\toprule
\multirow{2}{*}{Method}    &\multicolumn{4}{c|}{AP$_{3D} (\%)$ (\textit{validation set})}                                                             & \multicolumn{4}{c}{AP$_{3D} (\%)$ (\textit{test set})}                                                            \\

&                           \multicolumn{1}{c}{mAP}
                        &                           \multicolumn{1}{c}{Easy}           & \multicolumn{1}{c}{Mod.}           & \multicolumn{1}{c|}{Hard}           & \multicolumn{1}{c}{mAP}& \multicolumn{1}{c}{Easy}           & \multicolumn{1}{c}{Mod.}           & Hard           \\
                        \midrule
 Voxel R-CNN \upcite{voxelrcnn}                      & \multicolumn{1}{c}{86.84}   & \multicolumn{1}{c}{92.38}        & \multicolumn{1}{c}{85.29}          & 82.86          & \multicolumn{1}{c}{83.19}    & \multicolumn{1}{c}{90.90}      & \multicolumn{1}{c}{81.62}          & 77.06 \\
VFF \upcite{vff}                       & \multicolumn{1}{c}{86.91}   & \multicolumn{1}{c}{92.31}        & \multicolumn{1}{c}{85.51}          & 82.92          & \multicolumn{1}{c}{83.62}    & \multicolumn{1}{c}{89.50}      & \multicolumn{1}{c}{82.09}          & 79.29 \\
CAT-Det  \upcite{cat-det}                      & \multicolumn{1}{c}{83.58}   & \multicolumn{1}{c}{90.12}        & \multicolumn{1}{c}{81.46}          & 79.15          & \multicolumn{1}{c}{82.62}    & \multicolumn{1}{c}{89.87}      & \multicolumn{1}{c}{81.32}          & 76.68 \\
LoGoNet  \upcite{logonet}                                 & \multicolumn{1}{c}{87.13}  & \multicolumn{1}{c}{92.04}          & \multicolumn{1}{c}{85.04}          & \multicolumn{1}{c|}{84.31}         & \multicolumn{1}{c}{85.87}  & \multicolumn{1}{c}{91.80}          & \multicolumn{1}{c}{\textbf{85.06}}          & \multicolumn{1}{c}{80.74}       \\
Focals Conv-F \upcite{focalconv} & \multicolumn{1}{c}{-} & \multicolumn{1}{c}{-}        & \multicolumn{1}{c}{-}          & -          & \multicolumn{1}{c}{83.47} & \multicolumn{1}{c}{90.55}        & \multicolumn{1}{c}{82.28}          & 77.59    \\
SV-RCNN \upcite{SVRCNN} & \multicolumn{1}{c}{85.46} & \multicolumn{1}{c}{92.32}        & \multicolumn{1}{c}{83.24}          & 80.81          & \multicolumn{1}{c}{-} & \multicolumn{1}{c}{-}        & \multicolumn{1}{c}{-}          & -   \\
SID \upcite{wang2025boosting} & \multicolumn{1}{c}{87.87} & \multicolumn{1}{c}{92.87}        & \multicolumn{1}{c}{86.73}          & 84.01          & \multicolumn{1}{c}{-} & \multicolumn{1}{c}{-}        & \multicolumn{1}{c}{-}          & -   \\
CLEAN \upcite{zhang2025clean} & \multicolumn{1}{c}{80.36} & \multicolumn{1}{c}{88.80}        & \multicolumn{1}{c}{77.17}          & 75.12          & \multicolumn{1}{c}{-} & \multicolumn{1}{c}{-}        & \multicolumn{1}{c}{-}          & -   \\
Fade3D \upcite{ye2025fade3d} & \multicolumn{1}{c}{83.47} & \multicolumn{1}{c}{90.92}        & \multicolumn{1}{c}{82.00}          & 77.49         & \multicolumn{1}{c}{-} & \multicolumn{1}{c}{-}        & \multicolumn{1}{c}{-}          & -   \\
RAE3D \upcite{lian2025rae3d} & \multicolumn{1}{c}{83.16} & \multicolumn{1}{c}{91.68}
        & \multicolumn{1}{c}{80.31}          & 77.49         & \multicolumn{1}{c}{-} & \multicolumn{1}{c}{-}        & \multicolumn{1}{c}{-}          & -   \\

 \midrule
 RoboDistill(L)                              & \multicolumn{1}{c}{\textbf{89.03}}& \multicolumn{1}{c}{\textbf{93.55}} & \multicolumn{1}{c}{\textbf{88.11}} & \textbf{85.44}  & \multicolumn{1}{c}{\textbf{86.13}}& \multicolumn{1}{c}{\textbf{92.52}} & \multicolumn{1}{c}{84.65} & \textbf{81.21} \\
 RoboDistill(B)                              & \multicolumn{1}{c}{88.60}& \multicolumn{1}{c}{93.40} & \multicolumn{1}{c}{87.99} & 84.42 & \multicolumn{1}{c}{85.90}& \multicolumn{1}{c}{92.33} & \multicolumn{1}{c}{84.37} & 81.00\\
 RoboDistill(T)                               & \multicolumn{1}{c}{88.42}& \multicolumn{1}{c}{93.33} & \multicolumn{1}{c}{87.81} & 84.11 & \multicolumn{1}{c}{85.59}& \multicolumn{1}{c}{92.11} & \multicolumn{1}{c}{84.12} & 80.55  \\
\bottomrule
\end{tabular}}
\label{tab_kitti_val_test_val}
\par\vspace{2mm}
\caption{Comparison of RoboDistill (L/B/T) with representative multimodal state-of-the-art methods on the \textbf{nuScenes} validation and test sets in terms of NDS and mAP. RoboDistill is evaluated on an \textbf{NVIDIA A100 GPU} at an input resolution of $448 \times 800$ using FP16 precision, following the evaluation settings of DeepInteraction and TransFusion.}

\renewcommand\arraystretch{0.80}
\tabcolsep=3.89mm
\resizebox{\linewidth}{!}
{
\begin{tabular}{l|cc|cc|cc|cc}
    \toprule
\multirow{2}{*}{Method}  &     \multirow{2}{*}{LiDAR }  &\multirow{2}{*}{Camera }   &     \multicolumn{2}{c|}{\textit{validation set}} & \multicolumn{2}{c|}{\textit{test set}} & \multirow{2}{*}{Model size} & \multirow{2}{*}{FPS}  \\
 &&&NDS&mAP &NDS &mAP\\
\midrule

FUTR3D\upcite{chen2023futr3d} & VoxelNet & ResNet-101  & 68.3 & 64.5 & - & -&-&-\\
AutoAlignV2\upcite{autoalignv2} & VoxelNet & CSPNet & 71.2 & 67.1 & 72.4 & 68.4 &-&-\\
BEVFusion-mit\upcite{bevfusion-mit}& VoxelNet & Swin-T & 71.4 & 68.5 & 72.9 & 70.2 &-&-\\
DeepInteraction\upcite{deepinteraction} & VoxelNet & ResNet-50 & 72.6 & 69.9 & 73.4 & 70.8 &57.82M&4.9\\
CMT\upcite{cmt} & VoxelNet & ResNet-50 & 72.9 & 70.3 & 74.1 & 72.0&-&- \\
SparseFusion\upcite{sparsefusion}& VoxelNet & ResNet-50 & 72.8 & 70.4 & 73.8 & 72.0&-&- \\
TransFusion\upcite{transfusion} & VoxelNet & ResNet-50 & 71.3 & 67.5& 71.6 & 68.9&36.96M &6.2 \\
TiGDistill-BEV\upcite{xu2025tigdistill} & - & ResNet-101 & 52.0 & 41.2 &61.9 & 53.2&-&- \\
PARTNER\upcite{nie2026partner} & VoxelNet & Swin-T & 72.2 & 69.5& - & -&-&- \\
\midrule
RoboDistill(L) & VoxelNet & SAM & 72.7 & 70.5 & 73.4 & 71.3 &100.31M& 3.0\\
RoboDistill(B) & VoxelNet & FastSAM & 72.5& 70.2& 72.8 & 71.1 &84.42M &3.4\\
RoboDistill(T) & VoxelNet & MobileSAM & 72.3 & 70.1 & 72.5 & 70.8&15.23M &5.8 \\
\bottomrule
\end{tabular}
}
\label{tab_nuscenes_test_val}
\end{table}

\begin{table*}[t]
\caption{Robustness comparison across 27 OOD corruptions on the \textbf{KITTI-C} and \textbf{nuScenes-C} validation sets. The KITTI-C metric is car-class R$_{40}$ AP at moderate difficulty, whereas the nuScenes-C metric is mAP. All results are averaged over the official severity levels 1--5.}
\label{tab_kitti_c_car_moderate}
\renewcommand\arraystretch{0.80}
\tabcolsep=1.85mm
\resizebox{\linewidth}{!}
{
\begin{tabular}{ll|cccc|cccccccc}
\toprule
\multicolumn{2}{c|}{\multirow{3}{*}{\textbf{Corruptions}}}
& \multicolumn{4}{c|}{\textbf{KITTI-C}}
& \multicolumn{8}{c}{\textbf{nuScenes-C}} \\
&
& \multirow{1}{*}{RoboFusion}
& \multicolumn{3}{c|}{RoboDistill}
& \multirow{1}{*}{BEVFormer}
& \multirow{1}{*}{CenterPoint}
& \multirow{1}{*}{RoboFusion}
& \multicolumn{5}{c}{RoboDistill} \\
&&
& L & B & T
& & & & Camera only & LiDAR only & L & B & T \\
\midrule

\multicolumn{2}{c|}{\textbf{None}($\text{AP}_{\text{clean}}$)}
& 88.04 & \textbf{88.11} & 87.99 & 87.81
& 41.65 & 59.28 & 69.91 & 53.27 & 65.08 & \textbf{70.52} & 70.21 & 70.09 \\
\midrule

\multicolumn{1}{c|}{} & Snow
& 85.29 & \textbf{85.50} & 84.91 & 84.72
& 5.73 & 55.90 & 67.12 & 51.00 & 64.99 & \textbf{69.00} & 68.67 & 68.01 \\
\multicolumn{1}{c|}{} & Rain
& 86.48 & \textbf{86.81} & 86.51 & 86.22
& 24.97 & 56.08 & 67.58 & 50.87 & 64.03 & \textbf{68.77} & 68.65 & 68.50 \\
\multicolumn{1}{c|}{} & Fog
& 85.53 & \textbf{85.82} & 84.00 & 84.17
& 32.76 & 43.78 & 67.01 & 51.11 &63.91 & \textbf{68.00} & 67.81 & 67.61 \\
\multicolumn{1}{c|}{\multirow{-4}{*}{Weather}} & Sunlight
& 85.50 & \textbf{85.87} & 85.65 & 85.45
& 41.68 & 54.20 & 67.24 & 52.19 & 65.00 & \textbf{68.55} & 67.99 & 67.84 \\
\midrule

\multicolumn{1}{c|}{} & Density
& \textbf{85.71} & 85.33 & 85.12 & 84.78
& -& 58.60 & 69.48 & - & 63.83 & \textbf{70.21} & 69.92 & 69.74 \\
\multicolumn{1}{c|}{} & Cutout
& 83.17 & \textbf{84.20} & 81.30 & 81.21
&- & 56.28 & 69.18 & - & 62.99 & \textbf{70.22} & 69.87 & 69.52 \\
\multicolumn{1}{c|}{} & Crosstalk
& 84.12 & \textbf{85.87} & 85.45 & 84.97
&-& 56.64 & 68.68 & - & 63.12 & \textbf{69.20} & 68.87 & 68.76 \\
\multicolumn{1}{c|}{} & FOV loss
& - & - & - & -
& - & 20.84 & \underline{39.48} & - & 47.33 & \textbf{47.34} & 46.99 & 46.59 \\
\multicolumn{1}{c|}{} & Gaussian (L)
& \underline{76.56} & \textbf{81.12} & 80.81 & 79.98
& - & 45.79 & \underline{57.77} & - & 59.81 & \textbf{59.90} & 59.81 & 59.44 \\
\multicolumn{1}{c|}{} & Uniform (L)
& \underline{85.05} & \textbf{86.77} & 86.04 & 85.93
& - & 56.12 & \underline{64.57} & - & 64.22 & \textbf{67.00} & 66.85 & 66.02 \\
\multicolumn{1}{c|}{} & Impulse (L)
& \underline{85.26} & \textbf{87.32} & 87.00 & 86.46
& - & 57.67 & \underline{65.64} & - & 65.02 & \textbf{67.20} & 66.94 & 66.22 \\
\multicolumn{1}{c|}{} & Gaussian (C)
& \underline{82.16} & \textbf{84.08} & 83.93 & 83.64
& 15.04 & - & \underline{66.73} & 49.98 & - & \textbf{67.90} & 67.55 & 67.29 \\
\multicolumn{1}{c|}{} & Uniform (C)
& \underline{83.30} & \textbf{85.87} & 85.45 & 84.99
& 23.00 & - & \underline{65.77} & 48.88 & - & \textbf{67.99} & 67.34 & 66.89 \\
\multicolumn{1}{c|}{\multirow{-10}{*}{Sensor}} & Impulse (C)
& \underline{83.51} & \textbf{85.81} & 85.26 & 84.99
& 13.99 & - & \underline{64.82} & 47.90 & - & \textbf{67.21} & 66.99 & 66.81 \\
\midrule

\multicolumn{1}{c|}{} & Compensation
& \underline{41.88} & \textbf{48.34} & 47.00 & 46.93
& - & 11.02 & \underline{41.88} & - & 50.23 & \textbf{48.34} & 47.00 & 46.93 \\
\multicolumn{1}{c|}{} & Moving object
& \underline{49.30} & \textbf{53.22} & 52.14 & 51.90
& 20.22 & 44.30 & \underline{54.32} & 47.44 & 57.09 & \textbf{58.94} & 57.95 & 57.35 \\
\multicolumn{1}{c|}{\multirow{-3}{*}{Motion}} & Motion blur
& \underline{84.17} & \textbf{86.56} & 85.77 & 85.11
& 19.79 & - & \underline{67.21} & 52.23 & - & \textbf{68.91} & 68.37 & 68.31 \\
\midrule

\multicolumn{1}{c|}{\multirow{8}{*}{Object}} & Local density
& \underline{83.21} & \textbf{85.96} & 85.45 & 84.91
& - & 57.55 & \underline{66.74} & - & 64.34 & \textbf{67.46} & 67.21 & 67.10 \\
\multicolumn{1}{c|}{} & Local cutout
& \underline{77.22} & \textbf{78.78} & 77.93 & 77.56
& - & 48.36 & \underline{66.82} & - & 63.23 & \textbf{68.34} & 67.98 & 67.56 \\
\multicolumn{1}{c|}{} & Local Gaussian
& \underline{79.02} & \textbf{81.32} & 80.22 & 79.86
& - & 51.13 & \underline{65.08} & - & 64.02 & \textbf{68.00} & 67.87 & 67.61 \\
\multicolumn{1}{c|}{} & Local uniform
& \underline{84.69} & \textbf{86.87} & 86.12 & 85.77
& - & 57.87 & \underline{66.71} & - & 64.38 & \textbf{68.01} & 67.88 & 67.55 \\
\multicolumn{1}{c|}{} & Local impulse
& \underline{85.26} & \textbf{87.56} & 87.12 & 86.81
& - & 58.49 & \underline{66.53} & - & 64.42 & \textbf{68.02} & 67.90 & 67.79 \\
\multicolumn{1}{c|}{} & Shear
& \underline{55.42} & \textbf{61.32} & 60.21 & 59.23
& 24.71 &49.57 & \underline{62.33} & 48.23 & 64.03 & \textbf{64.10} & 64.02 & 63.84 \\
\multicolumn{1}{c|}{} & Scale
& \underline{74.23} & \textbf{76.21} & 75.92 & 74.32
& 17.64 & 51.13& \underline{65.47} & 52.01 & 64.01 & \textbf{64.99} & 64.82 & 64.39 \\
\multicolumn{1}{c|}{} & Rotation
& \underline{79.81} & \textbf{81.23} & 80.79 & 80.56
& 33.97 &54.68 & \underline{65.37} & 51.87 & 64.19 & \textbf{66.90} & 66.51 & 66.24 \\
\midrule

\multicolumn{1}{c|}{\multirow{2}{*}{Alignment}} & Spatial
& \underline{55.29} & \textbf{75.43} & 73.67 & 72.72
& - & - & \underline{58.49} & - & - & \textbf{68.21} & 67.48 & 67.52 \\
\multicolumn{1}{c|}{} & Temporal
& - & - & - & -
& - & - & \underline{40.93} & - & - & \textbf{59.87} & 59.46 & 58.92 \\
\bottomrule
\end{tabular}
}
\end{table*}

\begin{table}[t]
\centering
\caption{Reconstruction-loss matrix for RoboDistill (L) under different pretraining noise distributions and evaluation corruptions; lower values are better.}
\renewcommand\arraystretch{0.80}
\tabcolsep=2.99mm
\resizebox{\linewidth}{!}{

\begin{tabular}{l|cccccccc}
\toprule
\multirow{1}{*}{Pretraining noise}&\multicolumn{1}{c}{Gaussian (L)}           & \multicolumn{1}{c}{Uniform (L)}           & Impulse (L)           & \multicolumn{1}{c}{Gaussian (C)}& \multicolumn{1}{c}{Uniform (C) }           & \multicolumn{1}{c}{Impulse (C) }           & Density &  Crosstalk         \\
\midrule
Gaussian (L)
& \multicolumn{1}{c}{72.01}   & \multicolumn{1}{c}{63.77}        & \multicolumn{1}{c}{62.85}          & 55.08          & \multicolumn{1}{c}{54.33}    & \multicolumn{1}{c}{55.10}      & \multicolumn{1}{c}{54.22}          & 53.79 \\
Gaussian (C)
& 54.21 & 63.21& 62.88& 72.98& 62.70& 62.88& 64.43&63.91\\
Salt-and-pepper noise& 53.76& 60.34& 61.06& 62.05& 61.65& 65.63& 62.88&62.68
\\
Poisson noise& 53.09& 59.30& 58.90& 59.11& 58.56& 62.54& 62.65&64.87\\
\bottomrule
\end{tabular}}
\label{tab_ablation_noise_ood}
\end{table}

\begin{table}[t]
\centering
\caption{Effects of different SAM usage strategies on car-class R$_{40}$ AP on the \textbf{KITTI} and \textbf{KITTI-C} validation sets.}
\renewcommand\arraystretch{0.80}
\tabcolsep=7.99mm
\resizebox{\linewidth}{!}{

\begin{tabular}{l|cccc|cccc}
\toprule
\multirow{2}{*}{Solution}  & \multicolumn{4}{c|}{AP$_{3D} (\%)$}                                                             & \multicolumn{4}{c}{AP$_{Weather}(\%)$}                                                            \\ \cmidrule(r){2-9}
&                           \multicolumn{1}{c|}{mAP}
                        &                           \multicolumn{1}{c|}{Easy}           & \multicolumn{1}{c|}{Mod.}           & Hard           & \multicolumn{1}{c|}{Snow}& \multicolumn{1}{c|}{Rain}           & \multicolumn{1}{c|}{Fog}           & Sunlight           \\
                        \midrule
Offline
& \multicolumn{1}{c|}{80.41}   & \multicolumn{1}{c|}{88.45}        & \multicolumn{1}{c|}{77.12}          & 75.09          & \multicolumn{1}{c|}{-}    & \multicolumn{1}{c|}{-}      & \multicolumn{1}{c|}{-}          & - \\
Freeze
& \multicolumn{1}{c|}{86.45}   & \multicolumn{1}{c|}{91.90}        & \multicolumn{1}{c|}{84.83}          & 82.81          & \multicolumn{1}{c|}{45.22}    & \multicolumn{1}{c|}{47.81}      & \multicolumn{1}{c|}{63.21}          & 79.15 \\
Fine-tune
& \multicolumn{1}{c|}{88.00}   & \multicolumn{1}{c|}{92.76}        & \multicolumn{1}{c|}{86.99}          & 84.87          & \multicolumn{1}{c|}{58.00}    & \multicolumn{1}{c|}{56.76}      & \multicolumn{1}{c|}{69.32}          & 83.23 \\
\bottomrule
\end{tabular}}
\label{tab_abliation_offline_optim}
\end{table}

\begin{table*}[t]
\centering
\caption{Ablation results for SAM pretraining and the choice of wavelet basis in DGWA, together with a sensitivity analysis of the KD Fusion hyperparameters, on the \textbf{KITTI-C} validation set (car class, moderate difficulty, R$_{40}$ AP) and the \textbf{nuScenes-C} validation set (mAP).}
\renewcommand\arraystretch{0.82}
\scriptsize
\tabcolsep=0.25mm
\label{tab:pretrain_wavelet_kd}
\resizebox{\linewidth}{!}{
\begin{tabular}{ll|ccccccc|ccccccc}
\toprule
\multirow{2}{*}{Group} & \multirow{2}{*}{Setting} &
\multicolumn{7}{c|}{\textbf{KITTI-C validation}} &
\multicolumn{7}{c}{\textbf{nuScenes-C validation}} \\
& &
Snow & Rain & Fog & Sunlight & Density & Cutout & Crosstalk &
Snow & Rain & Fog & Sunlight & Density & Cutout & Crosstalk \\
\midrule

\multirow{2}{*}{Pretraining} & SAM
& 58.00& 56.76& 69.32& 83.23& 84.55& 83.61& 84.40
& 65.11& 66.21& 55.92& 57.90& 67.02& 65.44& 66.71 \\
& SAM-AD
& 85.50& 86.81& 85.82& 85.87& 85.33& 84.20& 85.87
& 69.00& 68.77& 68.00& 68.55& 70.21& 70.22& 69.20 \\
\midrule
\multirow{4}{*}{Wavelets}
& Daubechies-2
& 83.20& 83.42& 82.52& 81.92& 82.85& 81.20& 83.69
& 65.50& 64.38& 64.93& 65.13& 68.06& 67.35& 66.81 \\
& Symlets-4
& 82.98& 83.78& 84.00& 83.66& 81.43& 81.32& 81.95
& 66.28& 64.28& 66.48& 63.30& 67.60& 68.26& 67.55 \\
& Coiflets-1
& 81.86& 84.25& 83.67& 85.17& 79.65& 82.24& 81.90
& 67.91& 63.08& 66.62& 65.06& 66.72& 68.95& 66.01 \\

& \textbf{Haar}
& \textbf{85.50}& \textbf{86.81}& \textbf{85.82}& \textbf{85.87}& \textbf{85.33}& \textbf{84.20}& \textbf{85.87}
& \textbf{69.00}& \textbf{68.77}& \textbf{68.00}& \textbf{68.55}& \textbf{70.21}& \textbf{70.22}& \textbf{69.20} \\
\midrule
\multirow{7}{*}{Hyperparameters}
& $\alpha$=0.1, $\tau$=4
& 83.78& 84.94& 85.36& 84.64& 85.11& 83.09& 85.41
& 67.94& 66.85& 67.23& 67.56& 69.23& 68.89& 68.67 \\
& $\alpha$=0.3, $\tau$=4
& 83.54& 86.11& 85.55& 84.10& 84.72& 83.64& 85.60
& 68.51& 67.24& 66.58& 67.34& 69.47& 69.40& 68.42 \\

& \textbf{$\alpha$=0.5, $\tau$=4}
& \textbf{85.50}& \textbf{86.81}& \textbf{85.82}& \textbf{85.87}& \textbf{85.33}& \textbf{84.20}& \textbf{85.87}
& \textbf{69.00}& \textbf{68.77}& \textbf{68.00}& \textbf{68.55}& \textbf{70.21}& \textbf{70.22}& \textbf{69.20} \\
& $\alpha$=0.7, $\tau$=4
& 85.11& 86.70& 84.94& 83.98& 84.06& 82.73& 84.18
& 68.40& 66.80& 66.72& 67.18& 69.30& 68.43& 67.34 \\
& $\alpha$=0.5, $\tau$=1
& 85.36& 85.35& 84.60& 85.39& 83.98& 83.73& 85.29
& 67.01& 68.29& 67.34& 67.60& 69.67& 68.97& 67.25 \\
& $\alpha$=0.5, $\tau$=2
& 84.87& 86.65& 83.91& 84.97& 84.91& 82.32& 84.40
& 67.82& 67.32& 67.72& 66.87& 68.86& 68.49& 68.09 \\
& $\alpha$=0.5, $\tau$=8
& 84.45& 84.93& 84.73& 84.49& 84.94& 82.59& 85.44
& 68.62& 68.54& 67.00& 68.40& 69.27& 69.88& 68.58 \\
\bottomrule
\end{tabular}}
\end{table*}

\begin{table}[htp]
\centering
\caption{Ablation study of the contributions of different modules on the \textbf{KITTI-C} validation set (car class, moderate difficulty, R$_{40}$ AP) and the \textbf{nuScenes-C} validation set (mAP). DVCS denotes the \textbf{dynamically varying corruption setting}, a dynamic sensor-corruption evaluation protocol in which the corruption type (e.g., fog, rain, snow, occlusion, or sensor noise) and its severity evolve over time within each driving sequence to emulate rapidly changing real-world sensing conditions.}

\renewcommand\arraystretch{0.80}
\tabcolsep=2.09mm
\resizebox{\linewidth}{!}{
\begin{tabular}{c|cccc|ccccc|ccccc|c}
\toprule
\multirow{2}{*}{Method} &
\multirow{2}{*}{SAM-AD} &
\multirow{2}{*}{AD-FPN} &
\multirow{2}{*}{DGWA} &
\multirow{2}{*}{KD} &
\multicolumn{5}{c|}{\textbf{KITTI-C validation}} &
\multicolumn{5}{c|}{\textbf{nuScenes-C validation}} &
\multirow{2}{*}{FPS} \\
& & & & &
Snow & Rain & Fog & Sunlight &DVCS&
Snow & Rain & Fog & Sunlight &DVCS \\
\midrule
a) &  &  &  &  & 34.77 & 41.30 & 44.55 & 80.97&50.91 & 63.30 & 65.35 & 53.67 & 55.14& 59.97 & 10.8 \\
b) & \checkmark &  &  &  & 80.68 & 81.68 & 81.67 & 83.48&82.56 & 64.99 & 67.12 & 64.10 & 63.37 &65.21& 4.0 \\
c) & \checkmark & \checkmark &  &  & 82.32 & 83.60 & 82.39 & 83.98 & 83.41&66.88 & 68.04 & 64.02 & 65.87&66.82  & 3.6 \\
d) & \checkmark & \checkmark & \checkmark &  & 83.99 & 85.63 & 84.01 & 84.81 &84.91& 67.23 & 68.55 & 65.31 & 67.01 &67.22& 3.4 \\

e) & \checkmark & \checkmark & \checkmark & \checkmark & 85.50 & 86.81 & 85.82 & 85.87&86.10 & 69.00 & 68.77 & 68.00 & 68.55 &68.82& 3.0 \\
\bottomrule
\end{tabular}}
\label{tab_abliation_samad_all_modules}
\end{table}

\begin{figure*}[t]
    \centering
    \includegraphics[width=1\linewidth]{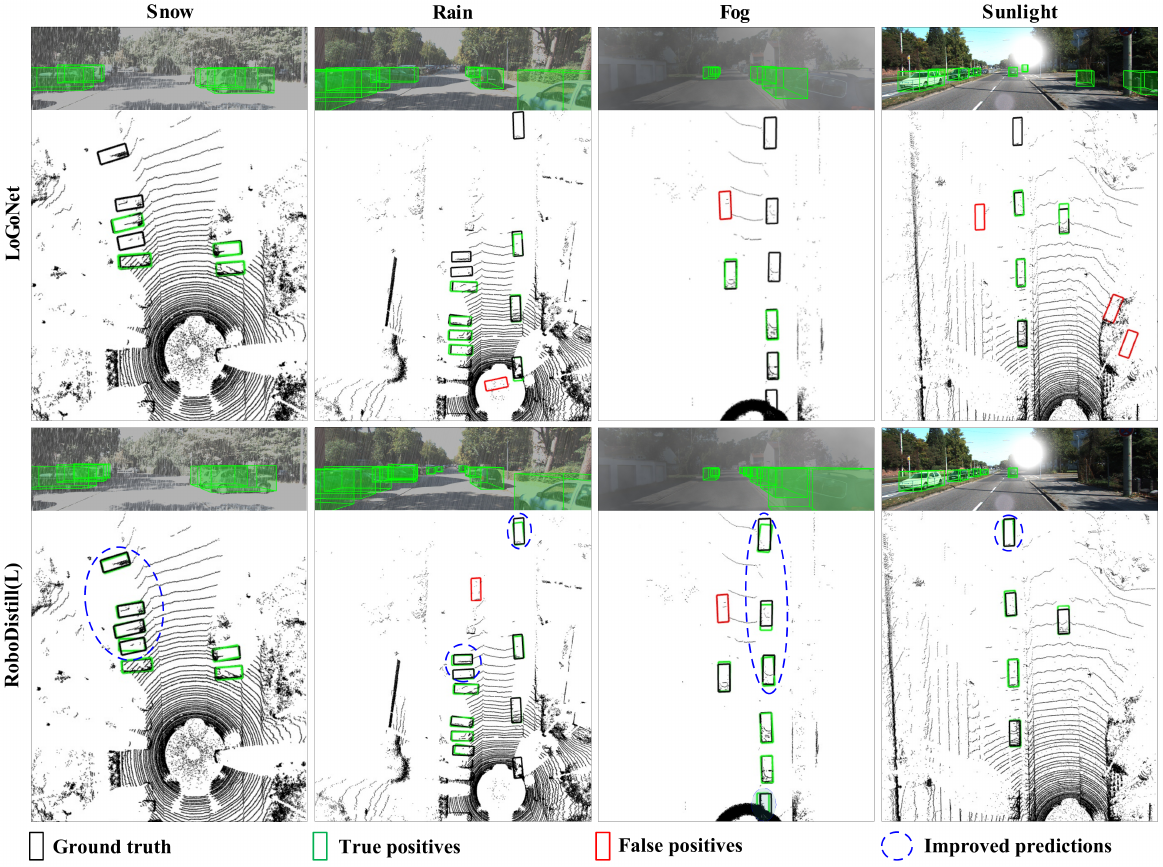}
    \caption{Visualization results for RoboDistill on KITTI-C. Red boxes denote false positives, green boxes denote true positives, and black boxes denote ground-truth annotations. Blue dashed ovals highlight the regions showing the most pronounced improvements.}
    \label{fig:vis-kittic}
\end{figure*}
\section{Experiments}
\subsection{Datasets}

\noindent \textbf{KITTI dataset.}
The KITTI dataset provides synchronized LiDAR point clouds and front-view camera images, comprising 3,712 training samples, 3,769 validation samples, and 7,518 test samples. The standard evaluation metric for object detection is mean average precision (mAP), computed over 40 recall positions (R40).

\noindent \textbf{nuScenes dataset.}
The nuScenes dataset is a large-scale 3D detection benchmark comprising 700 training scenes, 150 validation scenes, and 150 test scenes. Its data are captured by six multi-view cameras and a 32-channel LiDAR sensor, with 360-degree annotations for 10 object classes. The primary metrics for evaluating detection performance are mean average precision (mAP) and the nuScenes detection score (NDS).

\noindent \textbf{KITTI-C and nuScenes-C datasets.}
For robustness evaluation, Robustness3D \upcite{Robustness3d} defines 27 common corruptions for LiDAR and camera data to assess the corruption resistance of existing 3D detectors. It constructs corruption benchmarks \footnote{\url{https://github.com/thu-ml/3D_Corruptions_AD}}, including \textbf{KITTI-C} and \textbf{nuScenes-C}, by synthesizing corruptions on public datasets. Specifically, we use both \textbf{KITTI-C} and \textbf{nuScenes-C}. Robustness3D controls each corruption using officially defined \textbf{severity levels}, with corruption intensity selected from levels \textbf{1--5} rather than directly calibrated in physical units such as snowfall or visibility. Accordingly, all robustness experiments strictly follow the official protocol and report performance averaged over severity levels 1--5. Notably, Robustness3D \upcite{Robustness3d} adds noise only to the validation sets, while the training and test sets remain in their original clean state.

\subsection{Experimental Settings}
\noindent \textbf{Network architecture.}
Our RoboDistill framework comprises three variants: RoboDistill(L), RoboDistill(B), and RoboDistill(T), built on SAM-B \upcite{sam}, FastSAM \upcite{fastsam}, and MobileSAM \upcite{mobilesam}, respectively. The model design balances computational efficiency and feature extraction capacity for autonomous-driving tasks. Notably, the convolutional operations in FastSAM enable RoboDistill(B) to generate multiscale features, eliminating the need for the AD-FPN module.
We configure RoboDistill separately according to the evaluation metrics and characteristics of the KITTI and nuScenes datasets. For KITTI, we use Focals Conv \upcite{focalconv} as the baseline. The input voxel size is set to (0.05m, 0.05m, 0.1m), the vehicle anchor dimensions to [3.9, 1.6, 1.56], and the anchor rotations to [0, 1.57], matching the resolution and typical vehicle dimensions of KITTI. The data augmentation strategy follows Focals Conv-F \upcite{focalconv}.
For nuScenes, we use TransFusion \upcite{transfusion} as the baseline. The detection ranges along the X, Y, and Z axes are set to [-54m, 54m], [-54m, 54m], and [-5m, 3m], respectively, to accommodate the broader and more complex scenes in nuScenes. The input voxel size is set to (0.075m, 0.075m, 0.2m), with at most 10 point-cloud points per voxel.

\noindent \textbf{Training and testing details.}
The RoboDistill framework is trained with the Adam optimizer, and its image encoders are initialized with the pretrained weights of SAM, FastSAM, and MobileSAM, respectively. Training is conducted on 8 NVIDIA A100 GPUs to enable efficient training on the KITTI and nuScenes datasets. Inference time is also measured on an NVIDIA A100 GPU.
Specifically, for KITTI, RoboDistill uses a Focals Conv-based backbone \upcite{focalconv} and is trained for 80 epochs; for nuScenes, it uses TransFusion \upcite{transfusion} as the backbone and is trained for 20 epochs. During inference, we apply non-maximum suppression (NMS) in the region proposal network (RPN) with an IoU threshold of 0.7 to select the top 100 proposals, which are then passed to the detection head. After refinement by the detection head, NMS with an IoU threshold of 0.1 is applied again to remove redundant predictions, ensuring accurate and efficient detection.

\subsection{Evaluation Results}

\noindent \textbf{Results on KITTI.}
Table~\ref{tab_kitti_val_test_val} compares RoboDistill with existing methods on the KITTI validation and test sets. All three variants, L, B, and T, outperform the baseline across all evaluation difficulty levels. Compared with LoGoNet~\upcite{logonet}, the L variant improves $AP_{3D}$ at the hard difficulty level by 1.13 and 0.47 percentage points on the validation and test sets, reaching 85.44\% and 81.21\%, respectively. These results show that RoboDistill narrows the feature gap between images and point clouds while maintaining stable generalization.

\noindent \textbf{Results on nuScenes.}
The nuScenes evaluation results are reported in Table~\ref{tab_nuscenes_test_val}. The L variant achieves 72.7\% NDS and 70.5\% mAP on the validation set, and 73.4\% NDS and 71.3\% mAP on the test set. Compared with BEVFusion-mit, which uses a Transformer image branch, the L variant improves NDS and mAP by 1.3 and 2.0 percentage points on the validation set and by 0.5 and 1.1 percentage points on the test set, respectively. Its validation-set mAP also surpasses those of SparseFusion~\upcite{sparsefusion} and CMT~\upcite{cmt}. These results show that introducing a VFM, such as SAM, enhances feature representations and improves generalization and stability in complex scenes.

\noindent \textbf{Results on KITTI-C.}
Table~\ref{tab_kitti_c_car_moderate} compares the robustness of RoboDistill and RoboFusion on the KITTI-C validation set. Under weather corruptions (snow, rain, fog, and sunlight), RoboDistill(L) improves AP by 0.21, 0.33, 0.29, and 0.37 percentage points, respectively. Under sensor corruptions, except for Density, where it is 0.38 percentage points lower, it gains 1.03--4.56 percentage points on Cutout, Crosstalk, and Gaussian, uniform, and impulse noise. Under Compensation, Moving object, and Motion blur corruptions, the L variant reaches 48.34\%, 53.22\%, and 86.56\%, improving over RoboFusion by 6.46, 3.92, and 2.39 percentage points, respectively. It also maintains overall gains under object and spatial-alignment corruptions, further validating the effectiveness of KD Fusion for cross-modal alignment and fusion.

\noindent \textbf{Results on nuScenes-C.}
On the nuScenes-C validation set, Table~\ref{tab_kitti_c_car_moderate} shows that RoboDistill(L) maintains stable robustness. Under weather corruptions, its mAP reaches 69.00\%, 68.77\%, and 68.00\% for snow, rain, and fog, improving over RoboFusion by 1.88, 1.19, and 0.99 percentage points, respectively. RoboDistill(L) also achieves the best performance or substantial gains overall under sensor, motion, object, and alignment corruptions, demonstrating strong noise resistance and generalization.
We further conduct camera-only and LiDAR-only unimodal ablations to characterize the performance limits when one modality fails. The camera-only model obtains mAP values of 51.00\%, 50.87\%, and 51.11\% under snow, rain, and fog, respectively, substantially below the 64.99\%, 64.03\%, and 63.91\% of the LiDAR-only model and the 69.00\%, 68.77\%, and 68.00\% of the full model. This finding indicates that the camera branch is more sensitive to weather and imaging degradation. In contrast, the LiDAR-only model is more stable in most scenarios, but its performance under Compensation still drops from 65.08\% on clean data to 50.23\%. These results show that a single modality also has performance limits and that the fusion strategy can still be improved under certain corruptions.

\subsection{Ablation Studies}
\noindent \textbf{Reconstruction-loss comparison across noise distributions.}
Table~\ref{tab_ablation_noise_ood} compares the reconstruction losses of RoboDistill(L) under multiple noise distributions. Under \textit{Gaussian (C)}, \textit{salt-and-pepper noise}, and \textit{Poisson noise}, the \textit{Gaussian (L)} setting consistently achieves the lowest losses of 54.21, 53.76, and 53.09, respectively, demonstrating strong cross-distribution generalization. Overall, this experiment verifies the robustness of RoboDistill(L) to multiple types of OOD noise.

\noindent \textbf{Effects of different ways of using SAM.}
Table~\ref{tab_abliation_offline_optim} compares three ways of using SAM: \textit{Offline} (offline features), \textit{Freeze} (online but frozen), and \textit{Fine-tune} (end-to-end fine-tuning). \textit{Offline} performs worst, with an $AP_{3D}$ mAP of 80.41\% and Easy, Moderate, and Hard scores of 88.45\%, 77.12\%, and 75.09\%, respectively. \textit{Freeze} raises mAP to 86.45\% and reaches 45.22\%, 47.81\%, and 63.21\% under snow, rain, and fog corruptions on KITTI-C, respectively. \textit{Fine-tune} performs best, reaching an mAP of 88.00\%, with Easy, Moderate, and Hard scores of 92.76\%, 86.99\%, and 84.87\%, respectively; it reaches 58.00\% and 83.23\% under snow and sunlight corruptions. These results show that end-to-end optimization of SAM substantially improves generalization and robustness.

\noindent \textbf{Effect of SAM pretraining.}
Table~\ref{tab:pretrain_wavelet_kd} compares SAM and SAM-AD on KITTI-C under weather and sensor corruptions. Compared with the original SAM, SAM-AD pretrained on autonomous-driving data delivers substantial improvements across all corruptions, reaching 85.50\%/86.81\%/85.82\% under snow/rain/fog weather corruptions and 85.33\%/84.20\%/85.87\% under Density/Cutout/Crosstalk sensor corruptions. These results show that autonomous-driving-oriented pretraining substantially improves the robustness and generalization of SAM.

\noindent \textbf{Performance of different wavelets.}
Table~\ref{tab:pretrain_wavelet_kd} compares the effects of different wavelet bases in DGWA. Compared with Daubechies-2, Symlets-4, and Coiflets-1, our Haar wavelet achieves the best results on both KITTI-C and nuScenes-C, reaching 85.50/86.81/85.82 under Snow/Rain/Fog on KITTI-C. This result indicates that the selected wavelet basis more effectively suppresses noise while preserving critical structural information.

\noindent \textbf{Sensitivity to different hyperparameters.}
Table~\ref{tab:pretrain_wavelet_kd} reports a sensitivity analysis of the weight coefficient $\alpha$ and temperature coefficient $\tau$ in KD Fusion. Overall performance varies only slightly across settings, indicating that the method is insensitive to these hyperparameters. The setting $\alpha{=}0.5,\tau{=}4$ achieves the best and most stable results.

\noindent \textbf{Analysis of the RoboDistill modules.}
Table~\ref{tab_abliation_samad_all_modules} presents ablation results for the key modules of SAM-AD-based RoboDistill(L), including AD-FPN, DGWA, and KD Fusion. The results show that SAM-AD substantially improves the Focals Conv~\upcite{focalconv} baseline from (34.77\%, 41.30\%, 44.55\%, 80.97\%) to (80.68\%, 81.68\%, 81.67\%, 83.48\%). Adding AD-FPN, DGWA, and KD Fusion further yields consistent performance improvements, validating the importance of each module. Specifically, AD-FPN enhances multiscale feature representations, DGWA dynamically adjusts weights to optimize multimodal fusion, and KD Fusion transfers knowledge across modalities. Together, these modules improve the robustness and generalization of RoboDistill under OOD corruptions.

\subsection{Visualization}
Figure~\ref{fig:vis-kittic} presents comparative visualization results on the KITTI-C dataset for RoboDistill(L) and LoGoNet. Overall, compared with a state-of-the-art method such as LoGoNet~\upcite{logonet}, our method produces more accurate predictions, particularly for small and distant objects. By leveraging the generalization and robustness of visual foundation models, our method substantially improves the robustness of multimodal 3D object detection and effectively mitigates the effects of OOD corruptions in autonomous driving.

\section{Conclusion}
This paper proposes a robust and generalizable multimodal 3D object detection framework---\textbf{RoboDistill}---to address out-of-distribution (OOD) corruptions in autonomous-driving scenarios. By leveraging the strong generalization capability of visual foundation models (VFMs), such as SAM, RoboDistill effectively mitigates performance degradation caused by sensor noise, adverse weather, and environmental changes. To adapt VFMs to autonomous-driving tasks, we propose the domain-adaptive pretraining strategy \textbf{SAM-AD} and design the \textbf{AD-FPN} module to upsample and refine image features across multiple scales for seamless fusion with LiDAR features. The introduced \textbf{Depth-Guided Wavelet Attention (DGWA)} module effectively suppresses sensor noise, while the \textbf{multimodal fusion knowledge distillation (KD Fusion)} module further improves feature alignment and robustness by transferring high-quality knowledge from VFMs to the point-cloud network.
Compared with the previous version, RoboFusion, RoboDistill introduces substantial improvements in both architectural design and the knowledge-distillation mechanism and extends model generalization to 27 OOD corruption scenarios. Experiments on the KITTI-C and nuScenes-C benchmarks show that RoboDistill achieves stronger or competitive performance under most complex corruptions, demonstrating robust and scalable performance. Overall, RoboDistill effectively bridges the gap between clean-benchmark performance and real-world robustness, providing a reliable solution for multimodal 3D object detection in complex scenarios.

\noindent \textbf{Limitations and future work.}
First, RoboDistill relies heavily on the representation capability of visual foundation models. Although this substantially improves the generalization of the baseline model, it also increases model complexity. Second, because SAM and FastSAM incur substantial computational overhead, RoboDistill(L) and RoboDistill(B) have relatively low inference speeds; in contrast, RoboDistill(T) uses the more lightweight MobileSAM and achieves an inference speed closer to those of mainstream methods such as TransFusion. In future work, we will explore using SAM only during training to guide a lightweight student model, thereby further improving real-time performance. We will also investigate more complex and realistic corruption scenarios to further improve the robustness and practicality of the RoboDistill framework.

\clearpage
\normalsize
\pdfbookmark[0]{中文版本}{chinese-version}
\renewcommand{\figurename}{图}
\renewcommand{\tablename}{表}
\renewcommand{\refname}{参考文献}
\setcounter{section}{0}
\setcounter{subsection}{0}
\setcounter{figure}{0}
\setcounter{table}{0}
\setcounter{equation}{0}
\setcounter{footnote}{0}
\renewcommand*{\theHsection}{zh.\arabic{section}}
\renewcommand*{\theHsubsection}{zh.\arabic{section}.\arabic{subsection}}
\renewcommand*{\theHfigure}{zh.\arabic{figure}}
\renewcommand*{\theHtable}{zh.\arabic{table}}
\renewcommand*{\theHequation}{zh.\arabic{equation}}
\renewcommand*{\theHfootnote}{zh.\arabic{footnote}}

\begin{center}
{\large\bfseries 中文版本\par}
\vspace{1.2em}
{\LARGE\bfseries 基于视觉大模型的鲁棒多模态3D检测方法\par}
\vspace{1.2em}
{\large
宋子盈\textsuperscript{1,$\dagger$}\quad
刘林\textsuperscript{1,$\dagger$}\quad
潘虹宇\textsuperscript{2}\quad
徐少清\textsuperscript{3}\quad
杨磊\textsuperscript{4}\quad
郭明哲\textsuperscript{1}\quad
贾彩燕\textsuperscript{1,*}\par}
\vspace{0.8em}
{\small
\textsuperscript{1}北京交通大学计算机学院, 北京 100044\\
\textsuperscript{2}地平线机器人, 北京 100000\\
\textsuperscript{3}澳门大学, 澳门 999078\\
\textsuperscript{4}南洋理工大学, 新加坡 349562\\[0.3em]
\textsuperscript{$\dagger$}相同贡献.\quad
\textsuperscript{*}通讯作者: \texttt{cyjia@bjtu.edu.cn}}
\end{center}

\vspace{0.8em}
\noindent\textbf{摘要:}
多模态3D目标检测是实现自动驾驶系统鲁棒感知的关键任务, 因为它能够融合来自激光雷达和摄像头传感器的互补信息. 然而, 现有方法在分布外(OOD)噪声场景下(如传感器噪声、恶劣天气及环境变化)往往难以保持鲁棒性. 为解决这些问题, 我们提出了一种鲁棒且具备良好泛化能力的多模态3D检测框架---\textbf{RoboDistill}, 该框架充分利用视觉基础模型, 例如\textbf{Segment Anything Model (SAM)}, 以提升多模态3D目标检测性能.
首先, 我们提出了\textbf{SAM-AD}, 一种面向自动驾驶场景的领域自适应预训练策略, 通过对 SAM 进行特定领域微调, 充分挖掘其特征层面的语义信息.
其次, 我们设计了\textbf{AD 特征金字塔网络(AD-FPN)}, 对 SAM 提取的特征进行多尺度细化与上采样, 实现与激光雷达特征的无缝融合.
然后, 为应对传感器噪声的干扰, 我们提出了\textbf{深度引导小波注意力模块(DGWA)}, 该模块能够有效抑制高频噪声, 同时保留关键上下文信息.
最后, 我们引入\textbf{多模态融合知识蒸馏(KD Fusion)}机制, 其中预训练的 SAM-AD 作为教师网络, 将高质量的视觉知识蒸馏到轻量级点云网络中, 从而在噪声环境下显著提升鲁棒性.
大量实验结果表明, \textbf{RoboDistill} 在 27 种具有挑战性的分布外扰动下总体表现出更强或具有竞争力的性能与鲁棒性. 本研究有效弥合了视觉基础模型与3D目标检测之间的鸿沟, 为自动驾驶中的多模态感知提供了新的解决方案.

\medskip
\noindent\textbf{关键词:} 多模态融合, 知识蒸馏, 视觉基础模型, 3D目标检测, 自动驾驶

\medskip
\noindent\textbf{基金项目:} 国家自然科学基金, 批准号: 62536001(重点项目), 批准号: 62576026(面上项目)

\bigskip

\begin{figure}[!htbp]
\centering
\includegraphics[width=\textwidth]{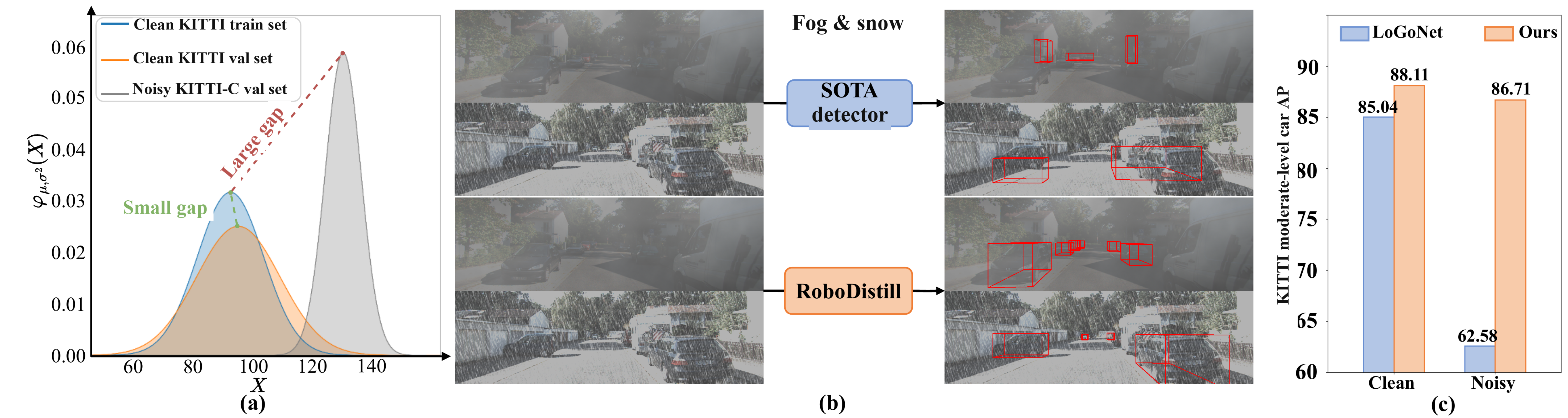}
\cnenfigcaption{(网络版彩图) \textbf{(a)} 我们用高斯分布刻画不同数据集的分布差异. 结果表明, OOD 噪声验证集与干净验证集之间存在显著分布鸿沟. 具体地, 横轴为数据集中图像平均像素值集合
\(X=\{x_i\}_{i=1}^{N}\),
其中
\(x_i=\frac{1}{H W 3}\sum_{h=1}^{H}\sum_{w=1}^{W}\sum_{c=1}^{3} I_{hwc}\),
\(N\) 为样本数, \(H,W\) 为图像尺寸, \(I_{hwc}\) 为像素值.
\textbf{(b)} VFMs(如 SAM \upcite{zh:sam}) 在多类噪声下具备较强鲁棒性, 但现有多模态3D检测方法在AD场景中仍易受OOD干扰.
\textbf{(c)} 因此, 我们提出 RoboDistill, 将 VFMs 引入SOTA多模态3D检测框架. 在图示的雾和雪扰动下, RoboDistill 的汽车类别中等难度 AP 为 86.71\%, 而 LoGoNet~\upcite{zh:logonet} 为 62.58\%, 即提高 24.13 个百分点; 在干净 KITTI \upcite{zh:kitti} 上亦保持更优表现.
}{(Color online) \textbf{(a)} We use Gaussian distributions to characterize distributional differences across datasets. The results reveal a clear gap between the OOD-corrupted and clean validation sets. Specifically, the x-axis represents the set of mean pixel intensities in a dataset,
\(X=\{x_i\}_{i=1}^{N}\),
where
\(x_i=\frac{1}{H W 3}\sum_{h=1}^{H}\sum_{w=1}^{W}\sum_{c=1}^{3} I_{hwc}\).
Here, \(N\) is the number of samples, \(H\) and \(W\) denote the image height and width, and \(I_{hwc}\) is the pixel value.
\textbf{(b)} Visual foundation models, such as SAM \upcite{zh:sam}, are robust to various corruptions; however, existing multimodal 3D detectors for autonomous driving remain vulnerable to OOD noise.
\textbf{(c)} RoboDistill integrates VFMs into a state-of-the-art multimodal 3D detection framework. Under the illustrated fog-and-snow corruption, RoboDistill achieves 86.71\% moderate-level car AP, compared with 62.58\% for LoGoNet~\upcite{zh:logonet}, an improvement of 24.13 percentage points, while also performing better on the clean KITTI \upcite{zh:kitti} dataset.
}
\label{zh:fig:motivation}
\end{figure}

\section{引言}
在自动驾驶场景中, 准确可靠的感知对于障碍物检测、目标跟踪和轨迹规划等任务至关重要 \upcite{zh:wang2023multi, zh:oza2023unsupervised}. 单模态(摄像头或激光雷达)系统在复杂环境下往往存在局限性. 基于摄像头的系统虽然提供丰富的语义信息, 但高度依赖光照条件并易受天气影响. 相比之下, 基于激光雷达的系统能够提供精准的几何和深度信息, 但数据稀疏且缺乏语义丰富性. 通过融合这两种模态, 多模态方法有望克服单一模态的局限性, 实现鲁棒的感知性能. 迄今为止, 多模态3D目标检测已成为鲁棒感知系统的基石, 因为它能够整合不同模态的互补信息 \upcite{zh:wang2023multi, zh:song2024robustness}.

多模态融合范式经历了显著发展, 从早期方法如 MV3D \upcite{zh:mv3d} 展示了多模态集成的可行性, 到后续方法如 PointPainting \upcite{zh:pointpainting} 用于语义增强, TransFusion \upcite{zh:transfusion} 用于跨模态交互, BEVFusion \upcite{zh:bevfusion-mit} 实现统一表示, 以及 CMT \upcite{zh:cmt} 利用隐式空间对齐. 尽管这些 SOTA 方法 \upcite{zh:song2023graphalign, zh:robofusion, zh:cmt, zh:bevfusion-mit, zh:GraphAlign_plus, zh:xushaoqing_fusionpating, zh:yin2024isfusion, zh:song2024contrastalign} 在某些方面实现了鲁棒性, 它们仍多在理想化条件下运行, 难以充分覆盖真实场景的复杂性. 例如, KITTI \upcite{zh:kitti} 数据集主要包含晴天场景, 缺乏雪、雾或大雨等多样天气条件, 且未考虑传感器噪声, 而这对于实际应用至关重要. 因此, KITTI-C \upcite{zh:Robustness3d} 被提出以模拟多种 OOD 噪声条件. 如图~\ref{zh:fig:motivation} 所示, OOD 噪声数据集与干净数据集之间存在明显的数据分布差异. 因此, 在干净数据集上训练的 SOTA 方法可能过拟合特定场景, 难以推广到夜间驾驶、极端天气或高传感器噪声等 OOD 环境, 这引出了一个关键研究问题: \textbf{如何在未知挑战性条件下保持方法的鲁棒性和可靠性?}

缩小 OOD 场景与干净训练数据集之间的差距可以通过提升数据集多样性以覆盖更多真实环境, 或设计对传感器噪声和环境变化鲁棒的算法. 然而, 构建覆盖各种环境条件的大规模数据集既困难又昂贵. 一个自然的途径是利用领域自适应(DA)技术, 该技术通常用于缩小源域与目标域之间的差距 \upcite{zh:wang2023ssda3d, zh:tsai2023viewer, zh:ST3D, zh:SPG}. 尽管 DA 技术可以通过减少对大量标注数据的依赖并将模型适配到新域来提升3D目标检测的鲁棒性, 它们仍存在固有局限性: 如处理显著域间差异困难, 标签分布变化带来的挑战, 以及过拟合风险 \upcite{zh:oza2023unsupervised}. 当源域与目标域差异较大时, DA 技术往往难以有效泛化, 导致目标域性能下降.

近年来, 自然语言处理与计算机视觉领域因基础模型的出现而发生变革性进展 \upcite{zh:sam, zh:gpt4, zh:fastsam, zh:mobilesam, zh:ma2024segment, zh:Sam_adapter}, 这些模型通过作为特征提取器、预测器或解释器, 在深度学习中建立了新范式 \upcite{zh:moor2023foundation, zh:fei2022towards}. 对于视觉基础模型而言, 它们展示了卓越的泛化能力, 归因于在大规模多样化数据集上的预训练 \upcite{zh:sam, zh:fastsam, zh:mobilesam}.
我们强调, 采用 SAM 并非依赖其分割输出本身, 而是利用其学习到的 ``目标区域一致性'' 与 ``边界结构先验''. 这些先验在雨雾模糊、局部遮挡和背景扰动下往往更稳定, 可提供高信噪比的语义锚点, 从而(1) 缓解图像侧噪声导致的语义漂移, (2) 促进2D--3D对齐并减少跨模态互信息损失, 最终提升OOD鲁棒性.
这些进展启发了利用 VFMs 来增强多模态3D目标检测鲁棒性的新方法, VFMs 可作为高层语义先验, 促进跨模态对齐, 缓解空间错位和特征冗余问题 \upcite{zh:seal}. 因此, 我们提出利用这些模型来应对 OOD 噪声场景下多模态3D目标检测系统面临的挑战.

因此, 本文提出鲁棒且具备良好泛化能力的框架---\textbf{RoboDistill}, 将视觉基础模型引入3D多模态检测流程, 旨在使现有检测器由干净、可控环境有效迁移至复杂噪声条件下的真实自动驾驶场景.
整体上, RoboDistill 形成 ``语义锚定 $\rightarrow$ 几何净化 $\rightarrow$ 受控迁移'' 的系统链路: SAM-AD 提供稳定语义先验, AD-FPN 将其对齐到可融合尺度, DGWA 在3D侧抑制噪声传播, KD Fusion 将可靠语义知识迁移到点云表征以提升鲁棒性.
首先, 考虑到通用 SAM 在自动驾驶场景中易受视角变化、动态目标与复杂背景等因素影响, 本文提出面向自动驾驶的域自适应预训练策略, 得到专用的 \textbf{SAM-AD}, 以对齐 VFM 能力与3D检测需求并增强跨模态特征对齐.
其次, 针对 VFM 特征在分辨率与层次结构上难以直接融入3D检测管线的问题, 本文设计 \textbf{AD-FPN} 对 SAM 特征进行高效上采样与细化, 实现高层语义与激光雷达特征的无缝融合, 同时兼顾空间分辨率与语义表达.
进一步地, 为抑制 OOD 场景下的噪声干扰, 本文提出 \textbf{深度引导小波注意力模块(DGWA)}: 对激光雷达深度特征进行小波分解, 将表示拆分为高频噪声与低频上下文分量, 并通过注意力机制选择性抑噪、保留关键结构信息, 从而在雾、雨、弱光等恶劣条件下获得更稳定的特征表征.
相较空间域滤波/归一化, 频域分解可将 ``非平稳噪声'' 与 ``结构化几何'' 在多尺度上解耦; 高频系数与噪声强度的相关性分析、频段消融(仅低频/仅高频/全频)以及空间域去噪基线对比共同支撑了该设计选择.
最后, 本文提出 \textbf{KD Fusion} 多模态融合知识蒸馏模块, 采用教师--学生范式, 将 \textbf{SAM-AD} 作为教师提供高质量图像表征并蒸馏至轻量级点云学生网络, 使其继承大规模预训练知识与领域泛化能力, 提升噪声与 OOD 条件下的特征提取与融合鲁棒性.
针对2D--3D域差异, 蒸馏损失仅在训练阶段用于从教师向学生传递指导, 且我们在 \emph{任务对齐的预测分布} 上蒸馏以避免特征空间强对齐带来的偏差; 同时引入教师置信度门控(高熵$\rightarrow$弱蒸馏)以抑制噪声场景下的不可靠软标签, 从而降低负迁移风险并保持稳定收益.
总之, VFMs 通过在大规模多样化数据集上的预训练, 展现出优异的泛化能力, 即使在挑战性环境下也能实现高效特征提取和鲁棒语义理解. 将 VFMs 集成到多模态3D检测流程中, RoboDistill 解决了跨模态对齐、噪声抑制和特征融合等关键问题, 为自动驾驶应用提供了鲁棒且可扩展的解决方案.

\section{相关工作}\label{zh:sec2}

\subsection{多模态3D目标检测}
多模态3D目标检测因能够利用来自不同传感器的互补信息而受到广泛关注, 尤其是在KITTI \upcite{zh:kitti} 和 nuScenes \upcite{zh:nuscenes} 等流行数据集上, 其中图像富含颜色和纹理等语义特征, 而LiDAR点云在捕获精确深度和几何结构方面表现优异. 现有研究 \upcite{zh:autoalignv2, zh:deepinteraction, zh:focalconv, zh:logonet, zh:sparsefusion, zh:song2023graphalign, zh:GraphAlign_plus} 强调跨模态特征的深度整合. 例如, PointPainting \upcite{zh:pointpainting} 利用预训练二维分割网络提取的语义特征增强LiDAR点云表示. 类似地, PointAugmenting \upcite{zh:pointaugmenting} 和 AutoAlignV2 \upcite{zh:autoalignv2} 提出更高效的跨模态交互策略, 利用全局图像特征提升点云处理性能. 这类方法虽提高了准确性, 但在真实噪声和环境变化下的鲁棒性仍是挑战. 近年来, BEV表示已成为多模态3D目标检测的主流范式. 基于BEV的方法 \upcite{zh:bevfusion-mit, zh:cmt, zh:song2025graphbev, zh:chen2023futr3d, zh:yin2024isfusion}, 如BEVFusion \upcite{zh:bevfusion-mit}, 将多模态特征统一到BEV空间, 实现高效融合并提升空间推理能力. 值得注意的是, GraphBEV \upcite{zh:song2025graphbev} 引入图结构的深度感知特征对齐, 以解决点云投影的不准确问题.
尽管现有SOTA多模态方法在干净、受控的数据集上表现出色, 但它们往往未能充分考虑真实世界场景中的复杂性, 如恶劣天气、传感器噪声和域偏移 \upcite{zh:song2024robustness, zh:Robustness3d}. 例如, KITTI和nuScenes等数据集在天气条件和环境因素上缺乏多样性, 导致在OOD场景下过拟合, 泛化能力下降.

\subsection{视觉基础模型在计算机视觉中的应用}
随着大型语言模型的快速发展 \upcite{zh:Flan-T5}, 计算机视觉领域涌现出众多视觉基础模型(Visual Foundation Models, VFMs) \upcite{zh:sam, zh:fastsam, zh:mobilesam, zh:XDecoder}. 这些VFMs通过在多样化数据集上的大规模预训练, 能够在新的视觉场景中展现出优异的泛化能力, 实现像素级特征提取. 然而, 大多数VFMs主要聚焦于二维视觉领域, 而将VFMs扩展到3D感知任务的研究仍然有限, 这为探索如何将现有二维VFMs适配或扩展到3D任务提供了研究空间. 首个VFM SAM \upcite{zh:sam} 基于Vision Transformer (ViT) \upcite{zh:vit} 构建, 并在包含1100万样本的SA-1B数据集上训练, 具备较强的场景泛化能力. FastSAM \upcite{zh:fastsam} 提供实时CNN方案, 大幅降低计算成本, 同时保持性能; MobileSAM \upcite{zh:mobilesam} 将SAM中的大型图像编码器(ViT-H)蒸馏为轻量级编码器, 兼容SAM的掩码解码器. 这些VFMs作为通用大模型, 为构建下游应用提供了强大工具. 尽管二维领域取得进展, VFMs在3D领域的探索仍处于起步阶段. SAM3D \upcite{zh:SAM3D} 作为纯LiDAR方法, 将3DLiDAR数据投影到二维BEV空间以利用SAM的泛化能力, 但模态交互有限导致性能欠佳. 与先前工作 RoboFusion 相比, 我们的 RoboDistill 利用 SAM 进行多模态知识蒸馏, 实现模型的知识迁移和学习.
综上所述, 现有SOTA多模态3D目标检测方法在OOD噪声场景中仍面临挑战, 难以弥合``干净''训练数据集与OOD噪声之间的差距;而VFMs所带来的泛化性与鲁棒性为视觉任务提供了新的机遇. 这启发我们利用VFMs的泛化能力与鲁棒性, 解决多模态3D检测中的OOD噪声泛化问题.

\section{RoboDistill}\label{zh:sec3}
\begin{figure}[!t]
\centering
\includegraphics[width=\textwidth]{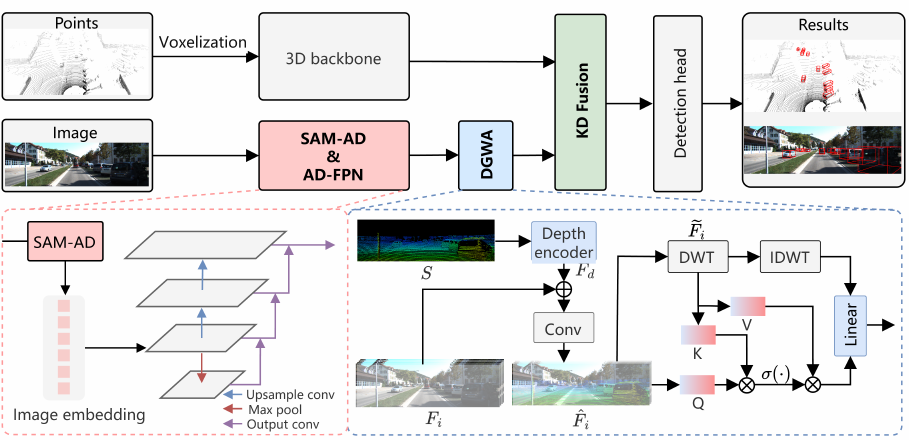}
\cnenfigcaption{(网络版彩图) RoboDistill 的总体框架.}{(Color online) Overall framework of RoboDistill.}
\label{zh:fig:framework}
\end{figure}

在本节中, 我们提出了\textbf{RoboDistill}框架,如图~\ref{zh:fig:framework}所示,  该方法在前期工作的基础上进行了扩展, 构建了一种全新的多模态策略, 以充分利用视觉基础模型(Visual Foundation Models, VFMs)的泛化能力. 与以往方法不同, \textbf{RoboDistill}通过结合VFMs与鲁棒的跨模态对齐技术, 显式地应对3D多模态目标检测中固有的分布外(Out-of-Distribution, OOD)噪声挑战, 为真实自动驾驶场景中的鲁棒性与可扩展性提供了新的解决方案. 值得注意的是,我们的detection head采用了Voxel RCNN \upcite{zh:voxelrcnn}一致的检测头.

\subsection{SAM-AD \& AD-FPN}
SAM~\upcite{zh:sam}作为一种代表性视觉基础模型, 凭借在大规模SA-1B数据集上的预训练展现出卓越的泛化能力. 该数据集包含超过1100万张样本和10亿个高质量掩码, 使SAM在多种视觉场景中表现出强大的鲁棒性与适应性. 目前, SAM系列模型~\upcite{zh:sam, zh:fastsam, zh:mobilesam}主要支持二维视觉任务. 然而, 直接将如SAM这类VFMs扩展至3D任务面临显著挑战, 这主要源于二维与3D任务在数据表示形式与任务需求上的根本差异. 为弥合这一差距, 本文将SAM与多模态3D模型相结合, 将二维鲁棒特征表示与3D点云特征融合, 从而实现更加鲁棒的融合特征表示.

\begin{figure}[!t]
\centering
\includegraphics[width=\textwidth]{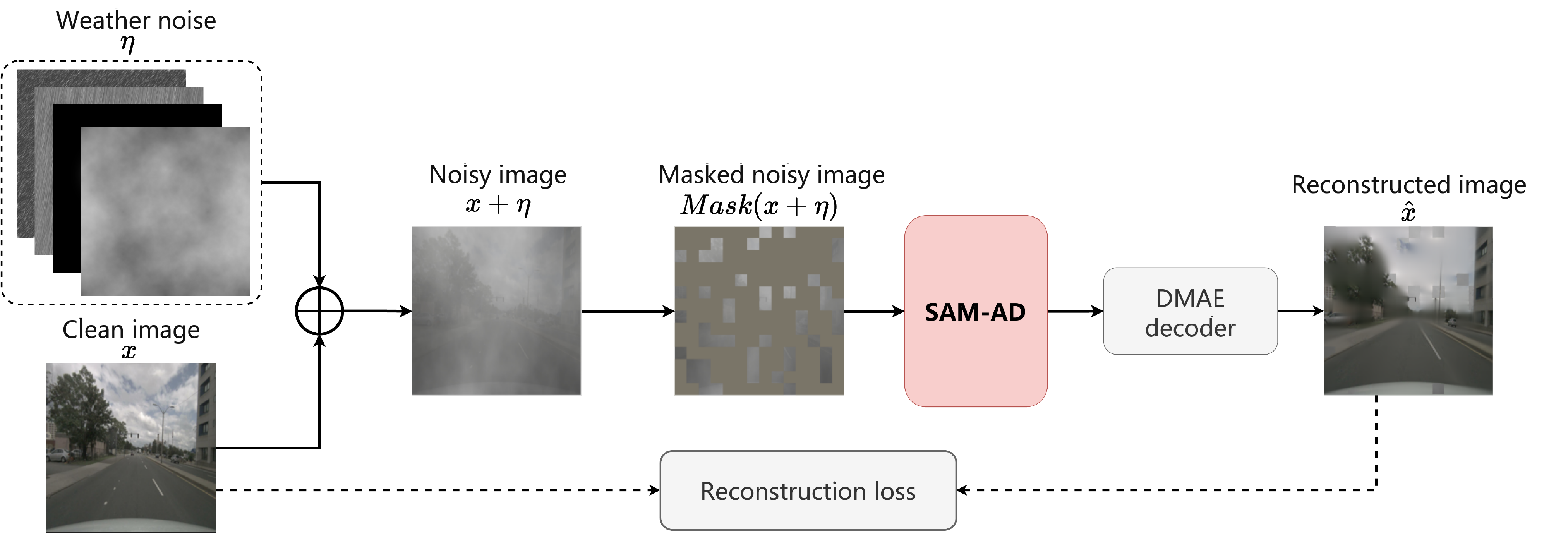}
\cnenfigcaption{(网络版彩图) 预训练框架示意图. 我们首先通过噪声扰动 $\eta$ 对干净图像 $x$ 进行扰动, 得到带噪图像 $x+\eta$. 随后, 在带噪图像上随机遮挡若干图像块, 生成掩码噪声图像 $Mask(x+\eta)$. 最后, 利用 SAM-AD 编码器和 DMAE 解码器从 $Mask(x+\eta)$ 重建对应的干净图像 $\hat{x}$.}{(Color online) Illustration of the pretraining framework. We corrupt a clean image $x$ with a noise perturbation $\eta$ to obtain $x+\eta$. We then randomly mask several patches in the corrupted image to obtain $Mask(x+\eta)$. The SAM-AD encoder and DMAE decoder are trained to reconstruct the corresponding clean image $\hat{x}$ from $Mask(x+\eta)$.}
\label{zh:fig:pretrain}
\end{figure}

\noindent \textbf{SAM-AD 模块. }
为使SAM更好地适应自动驾驶场景, 我们进行了领域自适应预训练, 得到针对自动驾驶任务优化的SAM-AD模型.具体而言, 我们从经典的KITTI~\upcite{zh:kitti}与nuScenes~\upcite{zh:nuscenes}等数据集中收集了大量自动驾驶图像, 涵盖多种驾驶场景与环境条件, 构建了基础AD图像数据集.在预训练过程中, 我们遵循DMAE框架~\upcite{zh:dmae}, 采用遮掩图像建模(Masked Image Modeling)对SAM与MobileSAM进行自监督预训练.具体地, 如图~\ref{zh:fig:pretrain}所示, 设$x$为从AD数据集中采样的干净图像, $\eta$为基于~\upcite{zh:Robustness3d}在$x$上施加扰动后得到的噪声图像集合.考虑到自动驾驶系统在真实部署中常面临复杂天气、光照变化以及成像退化等OOD干扰, 我们在预训练阶段采用了Robustness3D\upcite{zh:Robustness3d}中定义的全部噪声类型进行统一建模.例如,在nuScenes设置下, 共包含27种不同噪声类型, 且每种噪声均设置1--5级扰动强度, 以覆盖从轻度退化到重度退化的多种场景分布偏移.与针对单一噪声类型分别训练独立模型的方式不同, 我们采用统一的多噪声混合预训练策略, 训练一个共享的大模型, 从而学习更加通用且鲁棒的视觉表征.对于FastSAM, 我们使用YOLOv8在AD数据集上对其分割头进行预训练.为防止过拟合, 我们采用随机缩放与裁剪的数据增强策略, 掩码比例设为0.75.所有模型均在8块NVIDIA A100 GPU上训练400个epoch.模型性能通过重建精度与下游3D目标检测指标进行评估, 结果表明SAM-AD在OOD场景下具有良好的鲁棒性与泛化能力.

\noindent \textbf{AD-FPN 模块. }
作为一种可提示(promptable)的分割模型, SAM由三部分组成:图像编码器(image encoder)、提示编码器(prompt encoder)和掩码解码器(mask decoder). 其中提示编码器与掩码解码器主要用于语义分割任务, 而本文重点利用图像编码器提取鲁棒且高质量的图像特征嵌入. SAM采用基于ViT(Vision Transformer)~\upcite{zh:vit}的图像编码器, 其生成的高维低分辨率嵌入特征步幅为16(即scale=$1/16$). 然而, 这类特征缺乏多尺度表示能力, 不利于在多目标尺寸场景中进行精确定位与检测.

为解决这一问题, 我们设计了基于FPN~\upcite{zh:fpn}思想的AD-FPN模块, 用于增强ViT嵌入的多尺度表达能力. 具体而言, AD-FPN以ViT嵌入为输入, 构建一系列多尺度特征$F_s$, 其步幅$s \in \{32,16,8,4\}$. 每个特征图$F_s \in \mathbb{R}^{\frac{H}{s} \times \frac{W}{s} \times C_s}$通过自底向上的方式逐步细化. 与传统FPN不同, AD-FPN针对ViT嵌入进行了特定优化, 在保留其语义鲁棒性的同时提升空间分辨率与细粒度特征表示能力. 这一改进使模型能够更准确地检测小型、远距离或被遮挡的目标, 显著缓解了自动驾驶场景中常见的视觉挑战. 此外, AD-FPN的引入有效提升了多模态感知管线在恶劣条件(如低光照或高密度交通)下的鲁棒性与适应性, 为下游3D检测与感知任务提供了可靠且高效的特征支撑.

\subsection{深度引导小波注意力模块(DGWA)}

尽管 SAM-AD 与 SAM 模块能够提取具有鲁棒性的图像特征, 但二维与3D特征域之间仍存在显著差距. 在受损环境中, 由于摄像头缺乏几何先验, 其采集的图像往往会放大噪声, 从而导致特征的负迁移效应. 为此, 本文提出了 \textbf{深度引导小波注意力模块(DGWA)}, 通过``深度引导特征增强''和``小波域噪声抑制''两种关键操作来提升特征的鲁棒性.

\noindent \textbf{深度引导特征增强. }
该模块通过融合图像特征与深度特征, 将几何先验引入图像特征中. 具体而言, 图像特征表示为 \(F_i \in \mathbb{R}^{\frac{H}{4} \times \frac{W}{4} \times C_i}\), 深度特征表示为 \(F_d \in \mathbb{R}^{\frac{H}{4} \times \frac{W}{4} \times C_i}\). 其中, $F_d$ 由稀疏 LiDAR 深度图 \(S \in \mathbb{R}^{H \times W \times 2}\) 经深度编码器(Depth Encoder)处理获得, 该过程通过将点云投影至图像坐标系实现. 随后, 通过卷积融合操作生成深度引导特征 $\hat{F_i}=Conv(Concat\{F_i, F_d\})$, 使特征中嵌入空间与几何信息, 从而显著增强其抗噪能力.

\noindent \textbf{小波域噪声抑制. }  我们默认采用 Haar 小波作为 DGWA 的基础小波基,原因在于其分解形式简洁、计算开销低,且能够有效分离低频结构信息与高频噪声成分,更适合自动驾驶鲁棒感知中的实时特征增强任务.
对于增强后的特征 \(\hat{F_i}\), 采用 Haar 小波变换将其分解为四个子带:一个低频子带 \(\widetilde{f}_i^{LL} \in \mathbb{R}^{\frac{H}{8} \times \frac{W}{8} \times C_i}\) 和三个高频子带 \((\widetilde{f}_i^{LH}, \widetilde{f}_i^{HL}, \widetilde{f}_i^{HH}) \in \mathbb{R}^{\frac{H}{8} \times \frac{W}{8} \times C_i}\). 低频子带保留了粗粒度的上下文信息, 而高频子带则捕获细粒度的边缘与纹理特征, 从而便于识别噪声信号. 将这些子带拼接后得到小波特征 \(\widetilde{F_i} \in \mathbb{R}^{\frac{H}{8} \times \frac{W}{8} \times 4C_i}\), 并通过小波注意力机制 \(Att_{\omega}\) 进行特征加权. 该注意力机制在保留有效信息的同时, 有选择地抑制噪声, 其计算过程如下:
\begin{align}
F_{att} = Att_{\omega}(\hat{F_i}, \widetilde{F_i}) = \sigma\left(\frac{\hat{F_i}W^q (\widetilde{F_i}W^k)^T}{\sqrt{C_i}}\right) \widetilde{F_i}W^v.
\end{align}

最后, 通过逆小波变换(IDWT)对特征进行重构, 并与注意力输出进行融合, 以生成最终的鲁棒特征表示 \(F_{out}\):
\begin{align}
   F_{out} = MLP(\text{Concat}(F_{att}, \hat{F_i})).
\end{align}

该设计使得 \(F_{out}\) 能够在空间域与频率域中同时抑制冗余噪声并保留关键信息, 从而提升下游任务(如目标检测与跨模态特征融合)的鲁棒性与泛化能力.

\subsection{多模态知识蒸馏融合(KD Fusion)}
\label{zh:sec:fusion}

\begin{figure}[!t]
\centering
\includegraphics[width=0.7\textwidth]{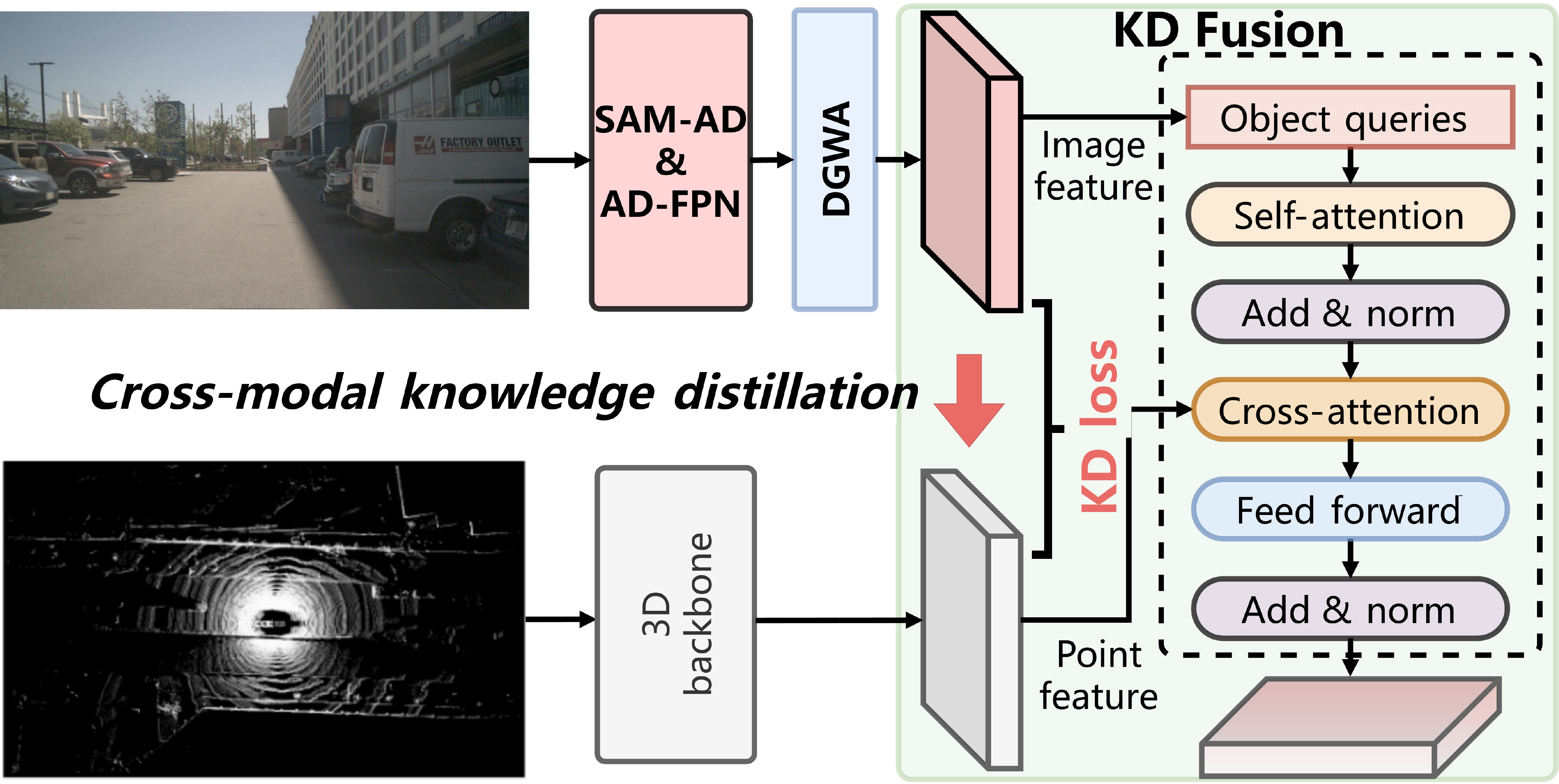}
\cnenfigcaption{(网络版彩图) \textbf{KD Fusion} 架构示意图. 图像分支采用 SAM-AD 作为教师模型, 点云分支作为学生模型. \textbf{KD Fusion} 将教师模型的知识迁移至学生模型, 并采用 Transformer 架构实现多模态特征对齐与融合.}{(Color online) Architecture of \textbf{KD Fusion}. The image branch uses SAM-AD as the teacher model, whereas the point cloud branch serves as the student model. \textbf{KD Fusion} transfers knowledge from the teacher to the student and uses a Transformer architecture for multimodal feature alignment and fusion.}
\label{zh:fig:kdfusion}
\end{figure}

尽管 SAM-AD 与 SAM 模块能够提取鲁棒的图像特征, 但在多模态融合中弥合二维与3D域之间的差距仍然是一个重大挑战. 在存在数据异质性的场景下, 直接对齐二维与3D模态往往难以保持互信息, 导致特征表达次优. 此外, 经过 SAM 处理的图像特征受益于其强大的泛化学习能力, 具有更高的鲁棒性与可迁移性. 为此, 本文将图像分支设定为 \textbf{教师模型(teacher)}, 点云分支设定为 \textbf{学生模型(student)}, 从而实现跨模态知识迁移以提升学生模型的学习能力. 本文提出的 \textbf{多模态融合知识蒸馏模块(KD Fusion)}, 如图~\ref{zh:fig:kdfusion} 所示, 通过教师-学生结构的互补性提升跨域鲁棒性.
在蒸馏过程中, 教师模型提供软目标(soft targets)与鲁棒先验; 在多模态推理过程中, 图像分支与点云分支共同参与特征融合.
该模块主要包括两个核心部分: 跨模态知识迁移与 Transformer 融合.

\noindent \textbf{跨模态知识迁移. }
教师模型 SAM-AD 从输入图像 \(I \in \mathbb{R}^{H \times W \times 3}\) 中提取丰富的语义信息, 并利用经过精炼的图像特征 $F_{out}$ 生成分类 logits \(T \in \mathbb{R}^{N \times C}\). 与此同时, 学生模型针对点云数据 \(P \in \mathbb{R}^{N \times D}\) 进行特征提取, 生成 logits \(S \in \mathbb{R}^{N \times C}\).
在优化层面, 我们首先完成 SAM-AD 的领域自适应预训练, 随后采用 ``\textbf{监督学习(CE) + 蒸馏学习(KL)}'' 的联合目标进行优化: CE 项保证任务精度下界, KL 项提供跨模态的类别关系与鲁棒先验, 从而在轻量化与性能保持之间取得平衡.

为缩小模态差距, 本文设计了一种知识蒸馏损失, 约束学生预测分布逼近教师预测分布, 同时保持任务精度. 考虑到二维语义空间与三维点云空间存在显著域差异, 我们选择在 \textbf{任务对齐的 logit/概率空间} 进行蒸馏(而非直接强制对齐中间特征), 以降低跨域不匹配带来的负迁移风险.
该损失函数由教师-学生预测分布之间的 Kullback--Leibler(KL) 散度与预测结果的交叉熵损失加权组合而成:
\begin{align}
\mathcal{L}_{KD}
=
\alpha \cdot \underbrace{w(T)\,\tau^{2}\,\mathrm{KL}\!\left(
\mathrm{softmax}\!\left(\tfrac{T}{\tau}\right)\, \big\| \,\mathrm{softmax}\!\left(\tfrac{S}{\tau}\right)
\right)}_{\text{distill from teacher}}
+ (1 - \alpha) \cdot \mathcal{L}_{CE}(S, Y),
\end{align}
其中, $\tau$ 表示温度系数, 用于平滑蒸馏分布以传递类间相似性; $\alpha$ 用于平衡蒸馏项与监督项, 在所有实验中我们固定采用 $\alpha=0.5$ 与 $\tau=4$, 以提升训练稳定性与可复现性.

此外, 为避免在严重噪声或OOD条件下发生负迁移(negative transfer), 我们引入教师置信度权重 $w(T)$ 对蒸馏强度进行自适应调节: 当教师预测不可靠(高熵/低置信)时减弱或关闭蒸馏, 从而抑制错误软标签对学生的误导. 一种简单实现为基于归一化熵的门控,
$w(T)=\max\!\left(0, 1-\tfrac{\mathcal{H}(\mathrm{softmax}(T))}{\log C}\right)$, 其中 $\mathcal{H}(\cdot)$ 为信息熵.

通过引入 KD Fusion, 学生模型能够在教师模型 SAM-AD 的指导下学习到更加鲁棒且紧凑的3D表示, 从而有效应对点云数据中的噪声与不确定性.

从定性角度看, SAM-AD 提供的指导主要体现为 ``语义先验 + 类间结构'' 两方面: (i) 语义先验来自大规模预训练与AD域自适应, 有助于在雾/雨/弱光等条件下维持稳定的类别判别线索; (ii) 类间结构通过温度平滑后的软分布传递, 能约束学生在噪声扰动下避免过度自信或决策边界漂移. 配合上述置信度门控, 教师仅在其 ``可靠区域'' 施加约束, 从而在跨域差异与噪声干扰并存时有效降低负迁移风险.

\noindent \textbf{Transformer 融合. }
为了实现图像与点云特征的高效融合, 本文在 KD Fusion 模块中引入基于 Transformer 的融合算子, 该算子结合了跨注意力机制与可学习的目标查询(object queries), 使点云网络(学生)能够通过跨模态知识蒸馏继承图像网络(教师)的鲁棒表示. 该融合算子主要包括三个关键过程:任务特定的自注意力优化、跨模态对齐注意力以及残差前馈精炼.

\begin{itemize}
    \item
    \textbf{自注意力优化(Self-attention for task-specific refinement). }
    融合模块利用可学习的目标查询(object queries)引导特征优化, 这些查询作为与目标任务(如目标检测)相关的嵌入向量. 首先, 自注意力机制通过建模查询之间的依赖关系来增强其对任务相关特征的关注能力, 其计算形式为:
    \begin{align}
    \text{Attention}(Q, K, V) = \text{Softmax}\left(\frac{QK^T}{\sqrt{d}}\right)V,
    \end{align}
    其中, \(Q, K, V\) 分别为由图像特征 $F_{out}$ 生成的查询、键和值矩阵. 该过程使目标查询能够自适应地学习任务特定的表征.

    \item
    \textbf{跨模态对齐注意力(Cross-attention for multimodal alignment). }
    在经过自注意力优化后, 目标查询进一步与图像与点云特征进行交互. 跨注意力机制将语义丰富的图像特征与空间细节充足的点云特征对齐, 实现模态间的互补信息融合. 该过程可表示为:
    \begin{align}
    \text{CrossAttention}(Q, K, V) = \text{Softmax}\left(\frac{QK^T}{\sqrt{d}}\right)V,
    \end{align}
    其中, \(Q\) 表示来自图像特征 $F_{out}$ 的查询向量, \(K, V\) 分别为由点云特征生成的键和值矩阵. 通过该机制, 学生模型(点云网络)能够有效继承教师模型(图像网络)的语义知识.

    \item
    \textbf{残差前馈精炼(Residual Feed-Forward Refinement). }
    为了稳定融合过程, 本文在每一层均引入残差连接与层归一化操作. 随后, 通过前馈神经网络(FFN)进一步提升融合特征的非线性表达能力, 从而增强下游任务的鲁棒性.
\end{itemize}

基于 Transformer 的融合算子能够同时整合来自图像与点云模态的语义、空间与几何信息. 通过跨注意力机制实现异构模态对齐, 该方法不仅促进了多模态特征的高效交互, 还通过知识蒸馏机制显著增强了模型在噪声或不完整输入条件下的鲁棒性.
\begin{table}[t]
\centering
\cnentablecaption{RoboDistill(L/B/T)与代表性最新方法在 \textbf{KITTI} 验证集与测试集上的 car 类 AP$_{3D}$(R$_{40}$)对比.}
{Comparison of RoboDistill (L/B/T) with representative state-of-the-art methods, including recent methods published since 2025, on the \textbf{KITTI} validation and test sets in terms of car-class AP$_{3D}$ (R$_{40}$).}
\renewcommand\arraystretch{0.80}
\tabcolsep=5.99mm
\resizebox{\linewidth}{!}
{
\begin{tabular}{l|cccc|cccc}
\toprule
\multirow{2}{*}{Method}    &\multicolumn{4}{c|}{AP$_{3D} (\%)$ (\textit{validation set})}                                                             & \multicolumn{4}{c}{AP$_{3D} (\%)$ (\textit{test set})}                                                            \\

&                           \multicolumn{1}{c}{mAP}
                        &                           \multicolumn{1}{c}{Easy}           & \multicolumn{1}{c}{Mod.}           & \multicolumn{1}{c|}{Hard}           & \multicolumn{1}{c}{mAP}& \multicolumn{1}{c}{Easy}           & \multicolumn{1}{c}{Mod.}           & Hard           \\
                        \midrule
 Voxel R-CNN \upcite{zh:voxelrcnn}                      & \multicolumn{1}{c}{86.84}   & \multicolumn{1}{c}{92.38}        & \multicolumn{1}{c}{85.29}          & 82.86          & \multicolumn{1}{c}{83.19}    & \multicolumn{1}{c}{90.90}      & \multicolumn{1}{c}{81.62}          & 77.06 \\
VFF \upcite{zh:vff}                       & \multicolumn{1}{c}{86.91}   & \multicolumn{1}{c}{92.31}        & \multicolumn{1}{c}{85.51}          & 82.92          & \multicolumn{1}{c}{83.62}    & \multicolumn{1}{c}{89.50}      & \multicolumn{1}{c}{82.09}          & 79.29 \\
CAT-Det  \upcite{zh:cat-det}                      & \multicolumn{1}{c}{83.58}   & \multicolumn{1}{c}{90.12}        & \multicolumn{1}{c}{81.46}          & 79.15          & \multicolumn{1}{c}{82.62}    & \multicolumn{1}{c}{89.87}      & \multicolumn{1}{c}{81.32}          & 76.68 \\
LoGoNet  \upcite{zh:logonet}                                 & \multicolumn{1}{c}{87.13}  & \multicolumn{1}{c}{92.04}          & \multicolumn{1}{c}{85.04}          & \multicolumn{1}{c|}{84.31}         & \multicolumn{1}{c}{85.87}  & \multicolumn{1}{c}{91.80}          & \multicolumn{1}{c}{\textbf{85.06}}          & \multicolumn{1}{c}{80.74}       \\
Focals Conv-F \upcite{zh:focalconv} & \multicolumn{1}{c}{-} & \multicolumn{1}{c}{-}        & \multicolumn{1}{c}{-}          & -          & \multicolumn{1}{c}{83.47} & \multicolumn{1}{c}{90.55}        & \multicolumn{1}{c}{82.28}          & 77.59    \\
SV-RCNN \upcite{zh:SVRCNN} & \multicolumn{1}{c}{85.46} & \multicolumn{1}{c}{92.32}        & \multicolumn{1}{c}{83.24}          & 80.81          & \multicolumn{1}{c}{-} & \multicolumn{1}{c}{-}        & \multicolumn{1}{c}{-}          & -   \\
SID \upcite{zh:wang2025boosting} & \multicolumn{1}{c}{87.87} & \multicolumn{1}{c}{92.87}        & \multicolumn{1}{c}{86.73}          & 84.01          & \multicolumn{1}{c}{-} & \multicolumn{1}{c}{-}        & \multicolumn{1}{c}{-}          & -   \\
CLEAN \upcite{zh:zhang2025clean} & \multicolumn{1}{c}{80.36} & \multicolumn{1}{c}{88.80}        & \multicolumn{1}{c}{77.17}          & 75.12          & \multicolumn{1}{c}{-} & \multicolumn{1}{c}{-}        & \multicolumn{1}{c}{-}          & -   \\
Fade3D \upcite{zh:ye2025fade3d} & \multicolumn{1}{c}{83.47} & \multicolumn{1}{c}{90.92}        & \multicolumn{1}{c}{82.00}          & 77.49         & \multicolumn{1}{c}{-} & \multicolumn{1}{c}{-}        & \multicolumn{1}{c}{-}          & -   \\
RAE3D \upcite{zh:lian2025rae3d} & \multicolumn{1}{c}{83.16} & \multicolumn{1}{c}{91.68}
        & \multicolumn{1}{c}{80.31}          & 77.49         & \multicolumn{1}{c}{-} & \multicolumn{1}{c}{-}        & \multicolumn{1}{c}{-}          & -   \\

 \midrule
 RoboDistill(L)                              & \multicolumn{1}{c}{\textbf{89.03}}& \multicolumn{1}{c}{\textbf{93.55}} & \multicolumn{1}{c}{\textbf{88.11}} & \textbf{85.44}  & \multicolumn{1}{c}{\textbf{86.13}}& \multicolumn{1}{c}{\textbf{92.52}} & \multicolumn{1}{c}{84.65} & \textbf{81.21} \\
 RoboDistill(B)                              & \multicolumn{1}{c}{88.60}& \multicolumn{1}{c}{93.40} & \multicolumn{1}{c}{87.99} & 84.42 & \multicolumn{1}{c}{85.90}& \multicolumn{1}{c}{92.33} & \multicolumn{1}{c}{84.37} & 81.00\\
 RoboDistill(T)                               & \multicolumn{1}{c}{88.42}& \multicolumn{1}{c}{93.33} & \multicolumn{1}{c}{87.81} & 84.11 & \multicolumn{1}{c}{85.59}& \multicolumn{1}{c}{92.11} & \multicolumn{1}{c}{84.12} & 80.55  \\
\bottomrule
\end{tabular}}
\label{zh:tab_kitti_val_test_val}
\par\vspace{2mm}
\cnentablecaption
{RoboDistill(L/B/T)与代表性多模态 SOTA 在 \textbf{nuScenes} 验证集与测试集上的对比:NDS/mAP. 我们在 \textbf{NVIDIA A100 GPU} 上进行评估, 输入分辨率设为 $448 \times 800$, 并采用 FP16 精度设置. 该评测配置与 DeepInteraction 和 TransFusion 保持一致.}
{Comparison of RoboDistill (L/B/T) with representative multimodal state-of-the-art methods on the \textbf{nuScenes} validation and test sets in terms of NDS and mAP. RoboDistill is evaluated on an \textbf{NVIDIA A100 GPU} at an input resolution of $448 \times 800$ using FP16 precision, following the evaluation settings of DeepInteraction and TransFusion.}

\renewcommand\arraystretch{0.80}
\tabcolsep=3.89mm
\resizebox{\linewidth}{!}
{
\begin{tabular}{l|cc|cc|cc|cc}
    \toprule
\multirow{2}{*}{Method}  &     \multirow{2}{*}{LiDAR }  &\multirow{2}{*}{Camera }   &     \multicolumn{2}{c|}{\textit{validation set}} & \multicolumn{2}{c|}{\textit{test set}} & \multirow{2}{*}{Model size} & \multirow{2}{*}{FPS}  \\
 &&&NDS&mAP &NDS &mAP\\
\midrule

FUTR3D\upcite{zh:chen2023futr3d} & VoxelNet & ResNet-101  & 68.3 & 64.5 & - & -&-&-\\
AutoAlignV2\upcite{zh:autoalignv2} & VoxelNet & CSPNet & 71.2 & 67.1 & 72.4 & 68.4 &-&-\\
BEVFusion-mit\upcite{zh:bevfusion-mit}& VoxelNet & Swin-T & 71.4 & 68.5 & 72.9 & 70.2 &-&-\\
DeepInteraction\upcite{zh:deepinteraction} & VoxelNet & ResNet-50 & 72.6 & 69.9 & 73.4 & 70.8 &57.82M&4.9\\
CMT\upcite{zh:cmt} & VoxelNet & ResNet-50 & 72.9 & 70.3 & 74.1 & 72.0&-&- \\
SparseFusion\upcite{zh:sparsefusion}& VoxelNet & ResNet-50 & 72.8 & 70.4 & 73.8 & 72.0&-&- \\
TransFusion\upcite{zh:transfusion} & VoxelNet & ResNet-50 & 71.3 & 67.5& 71.6 & 68.9&36.96M &6.2 \\
TiGDistill-BEV\upcite{zh:xu2025tigdistill} & - & ResNet-101 & 52.0 & 41.2 &61.9 & 53.2&-&- \\
PARTNER\upcite{zh:nie2026partner} & VoxelNet & Swin-T & 72.2 & 69.5& - & -&-&- \\
\midrule
RoboDistill(L) & VoxelNet & SAM & 72.7 & 70.5 & 73.4 & 71.3 &100.31M& 3.0\\
RoboDistill(B) & VoxelNet & FastSAM & 72.5& 70.2& 72.8 & 71.1 &84.42M &3.4\\
RoboDistill(T) & VoxelNet & MobileSAM & 72.3 & 70.1 & 72.5 & 70.8&15.23M &5.8 \\
\bottomrule
\end{tabular}
}
\label{zh:tab_nuscenes_test_val}
\end{table}

\begin{table*}[t]
\cnentablecaption
{在 \textbf{KITTI-C} 与 \textbf{nuScenes-C} 验证集上 27 种 OOD 扰动下的鲁棒性对比. KITTI-C 指标为 car 类 moderate 难度的 R$_{40}$ AP, nuScenes-C 指标为 mAP. 所有结果均为官方 severity level 1--5 下的平均值.}
{Robustness comparison across 27 OOD corruptions on the \textbf{KITTI-C} and \textbf{nuScenes-C} validation sets. The KITTI-C metric is car-class R$_{40}$ AP at moderate difficulty, whereas the nuScenes-C metric is mAP. All results are averaged over the official severity levels 1--5.}
\label{zh:tab_kitti_c_car_moderate}
\renewcommand\arraystretch{0.80}
\tabcolsep=1.85mm
\resizebox{\linewidth}{!}
{
\begin{tabular}{ll|cccc|cccccccc}
\toprule
\multicolumn{2}{c|}{\multirow{3}{*}{\textbf{Corruptions}}}
& \multicolumn{4}{c|}{\textbf{KITTI-C}}
& \multicolumn{8}{c}{\textbf{nuScenes-C}} \\
&
& \multirow{1}{*}{RoboFusion}
& \multicolumn{3}{c|}{RoboDistill}
& \multirow{1}{*}{BEVFormer}
& \multirow{1}{*}{CenterPoint}
& \multirow{1}{*}{RoboFusion}
& \multicolumn{5}{c}{RoboDistill} \\
&&
& L & B & T
& & & & Camera only & LiDAR only & L & B & T \\
\midrule

\multicolumn{2}{c|}{\textbf{None}($\text{AP}_{\text{clean}}$)}
& 88.04 & \textbf{88.11} & 87.99 & 87.81
& 41.65 & 59.28 & 69.91 & 53.27 & 65.08 & \textbf{70.52} & 70.21 & 70.09 \\
\midrule

\multicolumn{1}{c|}{} & Snow
& 85.29 & \textbf{85.50} & 84.91 & 84.72
& 5.73 & 55.90 & 67.12 & 51.00 & 64.99 & \textbf{69.00} & 68.67 & 68.01 \\
\multicolumn{1}{c|}{} & Rain
& 86.48 & \textbf{86.81} & 86.51 & 86.22
& 24.97 & 56.08 & 67.58 & 50.87 & 64.03 & \textbf{68.77} & 68.65 & 68.50 \\
\multicolumn{1}{c|}{} & Fog
& 85.53 & \textbf{85.82} & 84.00 & 84.17
& 32.76 & 43.78 & 67.01 & 51.11 &63.91 & \textbf{68.00} & 67.81 & 67.61 \\
\multicolumn{1}{c|}{\multirow{-4}{*}{Weather}} & Sunlight
& 85.50 & \textbf{85.87} & 85.65 & 85.45
& 41.68 & 54.20 & 67.24 & 52.19 & 65.00 & \textbf{68.55} & 67.99 & 67.84 \\
\midrule

\multicolumn{1}{c|}{} & Density
& \textbf{85.71} & 85.33 & 85.12 & 84.78
& -& 58.60 & 69.48 & - & 63.83 & \textbf{70.21} & 69.92 & 69.74 \\
\multicolumn{1}{c|}{} & Cutout
& 83.17 & \textbf{84.20} & 81.30 & 81.21
&- & 56.28 & 69.18 & - & 62.99 & \textbf{70.22} & 69.87 & 69.52 \\
\multicolumn{1}{c|}{} & Crosstalk
& 84.12 & \textbf{85.87} & 85.45 & 84.97
&-& 56.64 & 68.68 & - & 63.12 & \textbf{69.20} & 68.87 & 68.76 \\
\multicolumn{1}{c|}{} & FOV loss
& - & - & - & -
& - & 20.84 & \underline{39.48} & - & 47.33 & \textbf{47.34} & 46.99 & 46.59 \\
\multicolumn{1}{c|}{} & Gaussian (L)
& \underline{76.56} & \textbf{81.12} & 80.81 & 79.98
& - & 45.79 & \underline{57.77} & - & 59.81 & \textbf{59.90} & 59.81 & 59.44 \\
\multicolumn{1}{c|}{} & Uniform (L)
& \underline{85.05} & \textbf{86.77} & 86.04 & 85.93
& - & 56.12 & \underline{64.57} & - & 64.22 & \textbf{67.00} & 66.85 & 66.02 \\
\multicolumn{1}{c|}{} & Impulse (L)
& \underline{85.26} & \textbf{87.32} & 87.00 & 86.46
& - & 57.67 & \underline{65.64} & - & 65.02 & \textbf{67.20} & 66.94 & 66.22 \\
\multicolumn{1}{c|}{} & Gaussian (C)
& \underline{82.16} & \textbf{84.08} & 83.93 & 83.64
& 15.04 & - & \underline{66.73} & 49.98 & - & \textbf{67.90} & 67.55 & 67.29 \\
\multicolumn{1}{c|}{} & Uniform (C)
& \underline{83.30} & \textbf{85.87} & 85.45 & 84.99
& 23.00 & - & \underline{65.77} & 48.88 & - & \textbf{67.99} & 67.34 & 66.89 \\
\multicolumn{1}{c|}{\multirow{-10}{*}{Sensor}} & Impulse (C)
& \underline{83.51} & \textbf{85.81} & 85.26 & 84.99
& 13.99 & - & \underline{64.82} & 47.90 & - & \textbf{67.21} & 66.99 & 66.81 \\
\midrule

\multicolumn{1}{c|}{} & Compensation
& \underline{41.88} & \textbf{48.34} & 47.00 & 46.93
& - & 11.02 & \underline{41.88} & - & 50.23 & \textbf{48.34} & 47.00 & 46.93 \\
\multicolumn{1}{c|}{} & Moving object
& \underline{49.30} & \textbf{53.22} & 52.14 & 51.90
& 20.22 & 44.30 & \underline{54.32} & 47.44 & 57.09 & \textbf{58.94} & 57.95 & 57.35 \\
\multicolumn{1}{c|}{\multirow{-3}{*}{Motion}} & Motion blur
& \underline{84.17} & \textbf{86.56} & 85.77 & 85.11
& 19.79 & - & \underline{67.21} & 52.23 & - & \textbf{68.91} & 68.37 & 68.31 \\
\midrule

\multicolumn{1}{c|}{\multirow{8}{*}{Object}} & Local density
& \underline{83.21} & \textbf{85.96} & 85.45 & 84.91
& - & 57.55 & \underline{66.74} & - & 64.34 & \textbf{67.46} & 67.21 & 67.10 \\
\multicolumn{1}{c|}{} & Local cutout
& \underline{77.22} & \textbf{78.78} & 77.93 & 77.56
& - & 48.36 & \underline{66.82} & - & 63.23 & \textbf{68.34} & 67.98 & 67.56 \\
\multicolumn{1}{c|}{} & Local Gaussian
& \underline{79.02} & \textbf{81.32} & 80.22 & 79.86
& - & 51.13 & \underline{65.08} & - & 64.02 & \textbf{68.00} & 67.87 & 67.61 \\
\multicolumn{1}{c|}{} & Local uniform
& \underline{84.69} & \textbf{86.87} & 86.12 & 85.77
& - & 57.87 & \underline{66.71} & - & 64.38 & \textbf{68.01} & 67.88 & 67.55 \\
\multicolumn{1}{c|}{} & Local impulse
& \underline{85.26} & \textbf{87.56} & 87.12 & 86.81
& - & 58.49 & \underline{66.53} & - & 64.42 & \textbf{68.02} & 67.90 & 67.79 \\
\multicolumn{1}{c|}{} & Shear
& \underline{55.42} & \textbf{61.32} & 60.21 & 59.23
& 24.71 &49.57 & \underline{62.33} & 48.23 & 64.03 & \textbf{64.10} & 64.02 & 63.84 \\
\multicolumn{1}{c|}{} & Scale
& \underline{74.23} & \textbf{76.21} & 75.92 & 74.32
& 17.64 & 51.13& \underline{65.47} & 52.01 & 64.01 & \textbf{64.99} & 64.82 & 64.39 \\
\multicolumn{1}{c|}{} & Rotation
& \underline{79.81} & \textbf{81.23} & 80.79 & 80.56
& 33.97 &54.68 & \underline{65.37} & 51.87 & 64.19 & \textbf{66.90} & 66.51 & 66.24 \\
\midrule

\multicolumn{1}{c|}{\multirow{2}{*}{Alignment}} & Spatial
& \underline{55.29} & \textbf{75.43} & 73.67 & 72.72
& - & - & \underline{58.49} & - & - & \textbf{68.21} & 67.48 & 67.52 \\
\multicolumn{1}{c|}{} & Temporal
& - & - & - & -
& - & - & \underline{40.93} & - & - & \textbf{59.87} & 59.46 & 58.92 \\
\bottomrule
\end{tabular}
}
\end{table*}

\begin{table}[t]
\centering
\cnentablecaption{RoboDistill(L)在不同预训练噪声分布与评测扰动下的重建损失矩阵, 数值越低越好.}{Reconstruction-loss matrix for RoboDistill (L) under different pretraining noise distributions and evaluation corruptions; lower values are better.}
\renewcommand\arraystretch{0.80}
\tabcolsep=2.99mm
\resizebox{\linewidth}{!}{

\begin{tabular}{l|cccccccc}
\toprule
\multirow{1}{*}{Pretraining noise}&\multicolumn{1}{c}{Gaussian (L)}           & \multicolumn{1}{c}{Uniform (L)}           & Impulse (L)           & \multicolumn{1}{c}{Gaussian (C)}& \multicolumn{1}{c}{Uniform (C) }           & \multicolumn{1}{c}{Impulse (C) }           & Density &  Crosstalk         \\
\midrule
Gaussian (L)
& \multicolumn{1}{c}{72.01}   & \multicolumn{1}{c}{63.77}        & \multicolumn{1}{c}{62.85}          & 55.08          & \multicolumn{1}{c}{54.33}    & \multicolumn{1}{c}{55.10}      & \multicolumn{1}{c}{54.22}          & 53.79 \\
Gaussian (C)
& 54.21 & 63.21& 62.88& 72.98& 62.70& 62.88& 64.43&63.91\\
Salt-and-pepper noise& 53.76& 60.34& 61.06& 62.05& 61.65& 65.63& 62.88&62.68
\\
Poisson noise& 53.09& 59.30& 58.90& 59.11& 58.56& 62.54& 62.65&64.87\\
\bottomrule
\end{tabular}}
\label{zh:tab_ablation_noise_ood}
\end{table}

\begin{table}[t]
\centering
\cnentablecaption{不同 SAM 使用策略对 \textbf{KITTI} 与 \textbf{KITTI-C} 验证集汽车类别 R$_{40}$ AP 的影响.}{Effects of different SAM usage strategies on car-class R$_{40}$ AP on the \textbf{KITTI} and \textbf{KITTI-C} validation sets.}
\renewcommand\arraystretch{0.80}
\tabcolsep=7.99mm
\resizebox{\linewidth}{!}{

\begin{tabular}{l|cccc|cccc}
\toprule
\multirow{2}{*}{Solution}  & \multicolumn{4}{c|}{AP$_{3D} (\%)$}                                                             & \multicolumn{4}{c}{AP$_{Weather}(\%)$}                                                            \\ \cmidrule(r){2-9}
&                           \multicolumn{1}{c|}{mAP}
                        &                           \multicolumn{1}{c|}{Easy}           & \multicolumn{1}{c|}{Mod.}           & Hard           & \multicolumn{1}{c|}{Snow}& \multicolumn{1}{c|}{Rain}           & \multicolumn{1}{c|}{Fog}           & Sunlight           \\
                        \midrule
Offline
& \multicolumn{1}{c|}{80.41}   & \multicolumn{1}{c|}{88.45}        & \multicolumn{1}{c|}{77.12}          & 75.09          & \multicolumn{1}{c|}{-}    & \multicolumn{1}{c|}{-}      & \multicolumn{1}{c|}{-}          & - \\
Freeze
& \multicolumn{1}{c|}{86.45}   & \multicolumn{1}{c|}{91.90}        & \multicolumn{1}{c|}{84.83}          & 82.81          & \multicolumn{1}{c|}{45.22}    & \multicolumn{1}{c|}{47.81}      & \multicolumn{1}{c|}{63.21}          & 79.15 \\
Fine-tune
& \multicolumn{1}{c|}{88.00}   & \multicolumn{1}{c|}{92.76}        & \multicolumn{1}{c|}{86.99}          & 84.87          & \multicolumn{1}{c|}{58.00}    & \multicolumn{1}{c|}{56.76}      & \multicolumn{1}{c|}{69.32}          & 83.23 \\
\bottomrule
\end{tabular}}
\label{zh:tab_abliation_offline_optim}
\end{table}

\begin{table*}[t]
\centering
\cnentablecaption
{预训练、DGWA 小波基选择与 KD Fusion 超参数敏感性分析在 \textbf{KITTI-C 验证集}汽车类别(中等难度)R$_{40}$ AP 与 \textbf{nuScenes-C 验证集} mAP 上的汇总.}
{Ablation results for SAM pretraining and the choice of wavelet basis in DGWA, together with a sensitivity analysis of the KD Fusion hyperparameters, on the \textbf{KITTI-C} validation set (car class, moderate difficulty, R$_{40}$ AP) and the \textbf{nuScenes-C} validation set (mAP).}
\renewcommand\arraystretch{0.82}
\scriptsize
\tabcolsep=0.25mm
\label{zh:tab:pretrain_wavelet_kd}
\resizebox{\linewidth}{!}{
\begin{tabular}{ll|ccccccc|ccccccc}
\toprule
\multirow{2}{*}{Group} & \multirow{2}{*}{Setting} &
\multicolumn{7}{c|}{\textbf{KITTI-C validation}} &
\multicolumn{7}{c}{\textbf{nuScenes-C validation}} \\
& &
Snow & Rain & Fog & Sunlight & Density & Cutout & Crosstalk &
Snow & Rain & Fog & Sunlight & Density & Cutout & Crosstalk \\
\midrule

\multirow{2}{*}{Pretraining} & SAM
& 58.00& 56.76& 69.32& 83.23& 84.55& 83.61& 84.40
& 65.11& 66.21& 55.92& 57.90& 67.02& 65.44& 66.71 \\
& SAM-AD
& 85.50& 86.81& 85.82& 85.87& 85.33& 84.20& 85.87
& 69.00& 68.77& 68.00& 68.55& 70.21& 70.22& 69.20 \\
\midrule
\multirow{4}{*}{Wavelets}
& Daubechies-2
& 83.20& 83.42& 82.52& 81.92& 82.85& 81.20& 83.69
& 65.50& 64.38& 64.93& 65.13& 68.06& 67.35& 66.81 \\
& Symlets-4
& 82.98& 83.78& 84.00& 83.66& 81.43& 81.32& 81.95
& 66.28& 64.28& 66.48& 63.30& 67.60& 68.26& 67.55 \\
& Coiflets-1
& 81.86& 84.25& 83.67& 85.17& 79.65& 82.24& 81.90
& 67.91& 63.08& 66.62& 65.06& 66.72& 68.95& 66.01 \\

& \textbf{Haar}
& \textbf{85.50}& \textbf{86.81}& \textbf{85.82}& \textbf{85.87}& \textbf{85.33}& \textbf{84.20}& \textbf{85.87}
& \textbf{69.00}& \textbf{68.77}& \textbf{68.00}& \textbf{68.55}& \textbf{70.21}& \textbf{70.22}& \textbf{69.20} \\
\midrule
\multirow{7}{*}{Hyperparameters}
& $\alpha$=0.1, $\tau$=4
& 83.78& 84.94& 85.36& 84.64& 85.11& 83.09& 85.41
& 67.94& 66.85& 67.23& 67.56& 69.23& 68.89& 68.67 \\
& $\alpha$=0.3, $\tau$=4
& 83.54& 86.11& 85.55& 84.10& 84.72& 83.64& 85.60
& 68.51& 67.24& 66.58& 67.34& 69.47& 69.40& 68.42 \\

& \textbf{$\alpha$=0.5, $\tau$=4}
& \textbf{85.50}& \textbf{86.81}& \textbf{85.82}& \textbf{85.87}& \textbf{85.33}& \textbf{84.20}& \textbf{85.87}
& \textbf{69.00}& \textbf{68.77}& \textbf{68.00}& \textbf{68.55}& \textbf{70.21}& \textbf{70.22}& \textbf{69.20} \\
& $\alpha$=0.7, $\tau$=4
& 85.11& 86.70& 84.94& 83.98& 84.06& 82.73& 84.18
& 68.40& 66.80& 66.72& 67.18& 69.30& 68.43& 67.34 \\
& $\alpha$=0.5, $\tau$=1
& 85.36& 85.35& 84.60& 85.39& 83.98& 83.73& 85.29
& 67.01& 68.29& 67.34& 67.60& 69.67& 68.97& 67.25 \\
& $\alpha$=0.5, $\tau$=2
& 84.87& 86.65& 83.91& 84.97& 84.91& 82.32& 84.40
& 67.82& 67.32& 67.72& 66.87& 68.86& 68.49& 68.09 \\
& $\alpha$=0.5, $\tau$=8
& 84.45& 84.93& 84.73& 84.49& 84.94& 82.59& 85.44
& 68.62& 68.54& 67.00& 68.40& 69.27& 69.88& 68.58 \\
\bottomrule
\end{tabular}}
\end{table*}

\begin{table}[htp]
\centering
\cnentablecaption
{不同模块在 \textbf{KITTI-C 验证集}汽车类别(中等难度)R$_{40}$ AP 以及 \textbf{nuScenes-C 验证集} mAP 中的作用. DVCS 表示 \textbf{动态变化扰动设置}, 即一种动态传感器扰动评测协议: 在每段驾驶序列中, 扰动类型(如雾、雨、雪、遮挡或传感器噪声)及其强度随时间演化, 以模拟真实世界中快速变化的感知条件.
}
{Ablation study of the contributions of different modules on the \textbf{KITTI-C} validation set (car class, moderate difficulty, R$_{40}$ AP) and the \textbf{nuScenes-C} validation set (mAP). DVCS denotes the \textbf{dynamically varying corruption setting}, a dynamic sensor-corruption evaluation protocol in which the corruption type (e.g., fog, rain, snow, occlusion, or sensor noise) and its severity evolve over time within each driving sequence to emulate rapidly changing real-world sensing conditions.
}

\renewcommand\arraystretch{0.80}
\tabcolsep=2.09mm
\resizebox{\linewidth}{!}{
\begin{tabular}{c|cccc|ccccc|ccccc|c}
\toprule
\multirow{2}{*}{Method} &
\multirow{2}{*}{SAM-AD} &
\multirow{2}{*}{AD-FPN} &
\multirow{2}{*}{DGWA} &
\multirow{2}{*}{KD} &
\multicolumn{5}{c|}{\textbf{KITTI-C validation}} &
\multicolumn{5}{c|}{\textbf{nuScenes-C validation}} &
\multirow{2}{*}{FPS} \\
& & & & &
Snow & Rain & Fog & Sunlight &DVCS&
Snow & Rain & Fog & Sunlight &DVCS \\
\midrule
a) &  &  &  &  & 34.77 & 41.30 & 44.55 & 80.97&50.91 & 63.30 & 65.35 & 53.67 & 55.14& 59.97 & 10.8 \\
b) & \checkmark &  &  &  & 80.68 & 81.68 & 81.67 & 83.48&82.56 & 64.99 & 67.12 & 64.10 & 63.37 &65.21& 4.0 \\
c) & \checkmark & \checkmark &  &  & 82.32 & 83.60 & 82.39 & 83.98 & 83.41&66.88 & 68.04 & 64.02 & 65.87&66.82  & 3.6 \\
d) & \checkmark & \checkmark & \checkmark &  & 83.99 & 85.63 & 84.01 & 84.81 &84.91& 67.23 & 68.55 & 65.31 & 67.01 &67.22& 3.4 \\

e) & \checkmark & \checkmark & \checkmark & \checkmark & 85.50 & 86.81 & 85.82 & 85.87&86.10 & 69.00 & 68.77 & 68.00 & 68.55 &68.82& 3.0 \\
\bottomrule
\end{tabular}}
\label{zh:tab_abliation_samad_all_modules}
\end{table}

\begin{figure*}[t]
    \centering
    \includegraphics[width=1\linewidth]{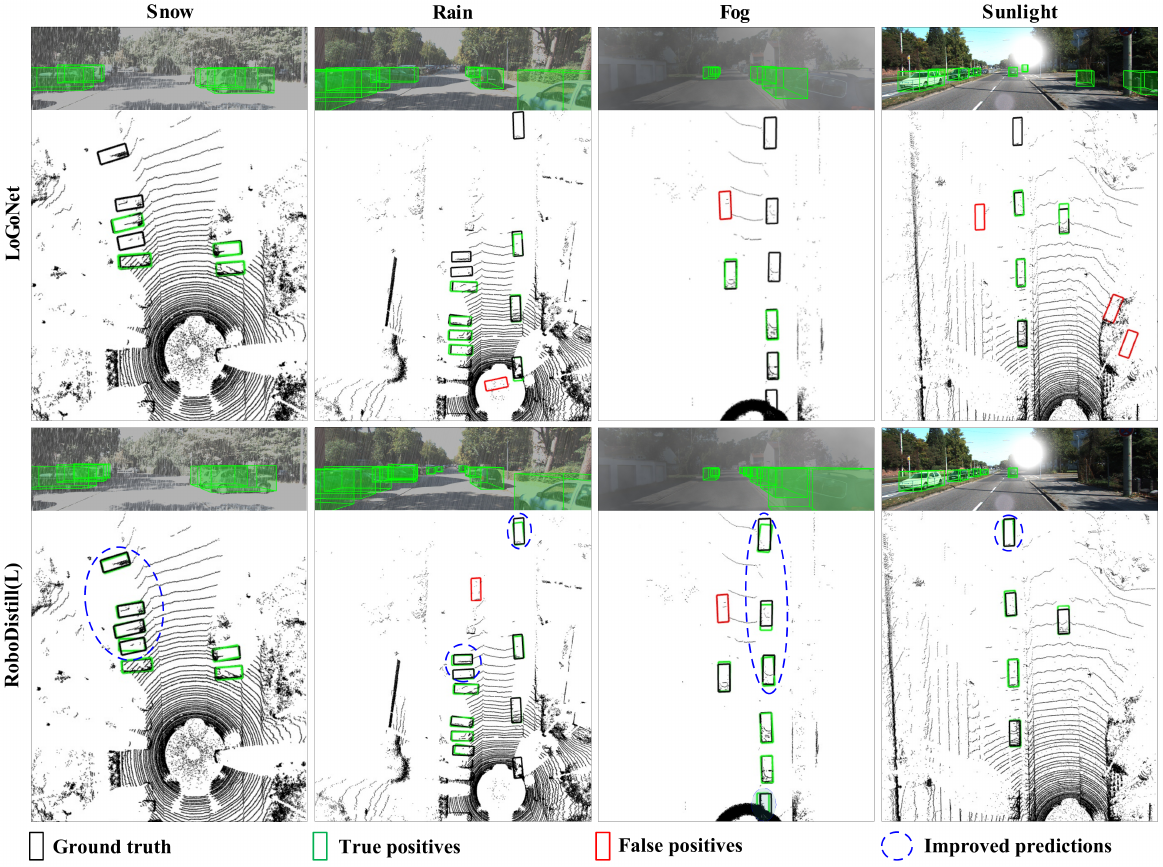}
    \cnenfigcaption{
        (网络版彩图) RoboDistill 在 KITTI-C 数据集上的可视化结果. 红色边界框表示误检, 绿色边界框表示正确检出, 黑色边界框表示真实标注. 蓝色虚线椭圆标出预测结果中提升最为显著的区域.
    }{
        (Color online) Visualization results for RoboDistill on KITTI-C. Red boxes denote false positives, green boxes denote true positives, and black boxes denote ground-truth annotations. Blue dashed ovals highlight the regions showing the most pronounced improvements.
    }
    \label{zh:fig:vis-kittic}
\end{figure*}
\section{实验}
\subsection{数据集}

\noindent \textbf{KITTI数据集.}
KITTI 数据集提供了同步采集的 LiDAR 点云和前视相机图像, 包含 3,712 个训练样本、3,769 个验证样本和 7,518 个测试样本. 目标检测的标准评估指标为平均精度均值(mAP), 采用 40 个召回位置(R40)计算.

\noindent \textbf{nuScenes数据集.}
nuScenes 数据集是一个大规模 3D 检测基准, 包含 700 个训练场景、150 个验证场景和 150 个测试场景. 数据通过六个多视角相机和 32 通道 LiDAR 传感器采集, 提供 10 类目标的 360 度标注. 评估检测性能的主要指标为平均精度均值(mAP)和 nuScenes 检测得分(NDS).

\noindent \textbf{KITTI-C 和 nuScenes-C数据集.}
在数据鲁棒性方面, 文献 \upcite{zh:Robustness3d} 为 LiDAR 和相机设计了 27 种常见扰动, 用于评测现有 3D 检测器的抗扰动能力.文献 \upcite{zh:Robustness3d} 构建了抗扰动基准 \footnote{\url{https://github.com/thu-ml/3D_Corruptions_AD}}, 包括 \textbf{KITTI-C} 和 \textbf{nuScenes-C}, 通过在公共数据集上合成扰动实现.具体而言, 本工作采用了 \textbf{KITTI-C} 和 \textbf{nuScenes-C}.需要说明的是, Robustness3D 对各类扰动采用官方定义的 \textbf{severity levels} 进行控制, 扰动强度从 \textbf{1--5} 级中选择, 而非以降雪量、能见度等物理单位直接标定.因此, 本文所有鲁棒性实验均严格遵循其官方设置, 并报告在 severity level 1--5 下的平均性能. 值得注意的是, 文献 \upcite{zh:Robustness3d} 仅对验证集添加了噪声, 训练集和测试集保持原始干净状态.

\subsection{实验设置}
\noindent \textbf{网络架构.}
我们的 RoboDistill 框架包含三个变体: RoboDistill(L)、RoboDistill(B) 和 RoboDistill(T), 分别基于 SAM-B \upcite{zh:sam}、FastSAM \upcite{zh:fastsam} 和 MobileSAM \upcite{zh:mobilesam} 构建. 模型设计兼顾计算效率与特征提取能力, 适配自动驾驶任务. 值得注意的是, FastSAM 中的卷积操作使得 RoboDistill(B) 能够生成多尺度特征, 从而无需 AD-FPN 模块.
根据 KITTI 和 nuScenes 数据集的不同评估指标及特性, 我们分别设置了 RoboDistill 的实验参数. 对于 KITTI 数据集, 我们以 Focals Conv \upcite{zh:focalconv} 为基线进行验证. 输入体素大小设置为 (0.05m, 0.05m, 0.1m), 车辆的 anchor 尺寸设置为 [3.9, 1.6, 1.56], anchor 旋转角度为 [0, 1.57], 以适应 KITTI 数据集的分辨率和车辆典型尺寸. 数据增强策略与 Focals Conv-F \upcite{zh:focalconv} 保持一致.
对于 nuScenes 数据集, 我们以 TransFusion \upcite{zh:transfusion} 为基线进行验证. 检测范围在 X、Y、Z 方向分别设置为 [-54m, 54m]、[-54m, 54m] 和 [-5m, 3m], 以适应 nuScenes 更广的场景覆盖和复杂性. 输入体素大小设置为 (0.075m, 0.075m, 0.2m), 每个体素内点云数量上限为 10.

\noindent \textbf{训练与测试细节.}
RoboDistill 框架采用 Adam 优化器训练, 并分别使用 SAM、FastSAM 和 MobileSAM 的预训练权重初始化图像编码器. 训练过程在 8 块 NVIDIA A100 GPU 上进行, 以确保在 KITTI 和 nuScenes 数据集上的高效训练. 同时, 评估推理时间也在 NVIDIA A100 GPU 上测量.
具体而言, 对于 KITTI 数据集, RoboDistill 使用基于 Focals Conv 的骨干网络 \upcite{zh:focalconv}, 训练 80 个 epoch; 对于 nuScenes 数据集, 使用 TransFusion \upcite{zh:transfusion} 作为骨干网络, 训练 20 个 epoch. 在模型推理阶段, 我们在区域候选网络(RPN)中采用非极大值抑制(NMS), IoU 阈值为 0.7, 以选择前 100 个候选区域, 并将其送入检测头. 经过检测头精炼后, 再次使用 IoU 阈值为 0.1 的 NMS 去除冗余预测, 以保证检测结果的精确性和效率.

\subsection{评估结果}

\noindent \textbf{KITTI 数据集结果.}
表~\ref{zh:tab_kitti_val_test_val} 给出了 RoboDistill 与现有方法在 KITTI 验证集和测试集上的对比结果. L、B 和 T 三种变体在各评估层级均优于基线. 与 LoGoNet~\upcite{zh:logonet} 相比, L 变体在 ``hard'' 难度下的 $AP_{3D}$ 在验证集和测试集上分别提升 1.13 和 0.47 个百分点, 达到 85.44\% 和 81.21\%. 结果表明, RoboDistill 能缩小图像与点云的特征差距, 并保持稳定的泛化表现.

\noindent \textbf{nuScenes 数据集结果.}
nuScenes 的评估结果见表~\ref{zh:tab_nuscenes_test_val}. L 变体在验证集取得 72.7\% NDS 和 70.5\% mAP, 在测试集取得 73.4\% NDS 和 71.3\% mAP. 与使用 Transformer 图像分支的 BEVFusion-mit 相比, L 变体在验证集上的 NDS 和 mAP 分别提升 1.3 和 2.0 个百分点, 在测试集上分别提升 0.5 和 1.1 个百分点. 其验证集 mAP 也优于 SparseFusion~\upcite{zh:sparsefusion} 与 CMT~\upcite{zh:cmt}. 结果表明, 引入 VFM(如 SAM)能增强特征表征, 提升复杂场景下的泛化与稳定性.

\noindent \textbf{KITTI-C 数据集结果.}
我们在表~\ref{zh:tab_kitti_c_car_moderate} 比较了 RoboDistill 与 RoboFusion 在 KITTI-C 验证集上的鲁棒性. 在天气扰动(雪、雨、雾、强光)下, RoboDistill(L) 的 AP 分别提升 0.21、0.33、0.29 和 0.37 个百分点. 在传感器扰动下, 除 Density 低 0.38 个百分点外, Cutout、Crosstalk 及高斯、均匀和脉冲噪声上的增益为 1.03--4.56 个百分点. 在 Compensation、Moving object 和 Motion blur 扰动下, L 变体分别达到 48.34\%、53.22\% 和 86.56\%, 比 RoboFusion 提升 6.46、3.92 和 2.39 个百分点. 在目标及空间对齐扰动下也整体保持增益, 进一步验证了 KD Fusion 在跨模态对齐与融合中的有效性.

\noindent \textbf{nuScenes-C 数据集结果.}
在 nuScenes-C 验证集上, 表~\ref{zh:tab_kitti_c_car_moderate} 显示 RoboDistill(L) 具有稳定鲁棒性. 在天气扰动下, 雪、雨和雾场景的 mAP 分别达到 69.00\%、68.77\% 和 68.00\%, 相比 RoboFusion 分别提升 1.88、1.19 和 0.99 个百分点. 在传感器、运动、目标及对齐扰动下, RoboDistill(L) 也整体取得最优或显著增益, 体现出较强的抗噪能力与泛化性能.
我们进一步开展了 camera-only 与 LiDAR-only 单模态消融分析, 以讨论模型在单模态失效时的性能边界. 结果表明, camera-only 在雪、雨和雾扰动下的 mAP 分别为 51.00\%、50.87\% 和 51.11\%, 明显低于 LiDAR-only 的 64.99\%、64.03\% 和 63.91\%, 也低于完整模型的 69.00\%、68.77\% 和 68.00\%, 说明相机分支对天气与成像退化更为敏感. 相比之下, LiDAR-only 在多数场景下更稳定, 但在 Compensation 扰动下仍由干净数据上的 65.08\% 降至 50.23\%. 该结果表明单一模态同样存在性能边界, 且个别扰动下的融合策略仍有改进空间.

\subsection{消融结果}
\noindent \textbf{在不同噪声分布下的重建损失对比实验.}
我们在表~\ref{zh:tab_ablation_noise_ood} 比较了 RoboDistill(L) 在多种噪声分布下的重建损失. 结果表明, 在 \textit{Gaussian (C)}、\textit{salt-and-pepper noise} 和 \textit{Poisson noise} 下, \textit{Gaussian (L)} 设置均取得最低损失, 分别为 54.21、53.76 和 53.09, 体现出较强的跨分布泛化能力. 总体而言, 该实验验证了 RoboDistill(L) 对多类 OOD 噪声具有较好的鲁棒性.

\noindent \textbf{不同 SAM 使用方式的影响.}
我们在表~\ref{zh:tab_abliation_offline_optim} 比较了 SAM 的三种使用方式: \textit{Offline}(离线特征)、\textit{Freeze}(在线但冻结)与 \textit{Fine-tune}(端到端微调). \textit{Offline} 性能最低, $AP_{3D}$ mAP 为 80.41\%, Easy、Moderate 和 Hard 分别为 88.45\%、77.12\% 和 75.09\%. \textit{Freeze} 将 mAP 提升至 86.45\%, 并在 KITTI-C 的雪、雨和雾扰动下分别达到 45.22\%、47.81\% 和 63.21\%. \textit{Fine-tune} 最优, mAP 达 88.00\%, Easy、Moderate 和 Hard 分别为 92.76\%、86.99\% 和 84.87\%; 雪和强光扰动下分别达到 58.00\% 和 83.23\%. 结果表明, 端到端优化 SAM 能显著增强泛化与鲁棒性.

\noindent \textbf{SAM 预训练的影响.}
我们在表~\ref{zh:tab:pretrain_wavelet_kd} 对比了 KITTI-C 上 SAM 与 SAM-AD 在天气与传感器扰动下的表现. 相比原始 SAM, 基于自动驾驶数据预训练的 SAM-AD 在各类扰动下均显著提升, 天气扰动中雪/雨/雾分别达到 85.50\%/86.81\%/85.82\%, 传感器扰动中 Density/Cutout/Crosstalk 分别为 85.33\%/84.20\%/85.87\%. 结果表明, 面向自动驾驶的预训练可显著增强 SAM 的鲁棒性与泛化能力.

\noindent \textbf{不同小波的性能.}
我们在表~\ref{zh:tab:pretrain_wavelet_kd} 对比了 DGWA 中不同小波基的影响. 相比 Daubechies-2、Symlets-4 与 Coiflets-1, 我们的 Haar 在 KITTI-C 与 nuScenes-C 上均取得最优结果, 在 KITTI-C 的 Snow/Rain/Fog 下分别达到 85.50/86.81/85.82, 表明所选小波基可更有效抑制噪声并保留关键结构信息.

\noindent \textbf{不同超参数的敏感性实验.}
我们在表~\ref{zh:tab:pretrain_wavelet_kd} 给出了 KD Fusion 中权重系数 $\alpha$ 与温度系数 $\tau$ 的敏感性分析. 不同设置下整体性能波动较小, 表明方法对超参数不敏感. 其中, $\alpha{=}0.5,\tau{=}4$ 取得最优且最稳定的结果.

\noindent \textbf{RoboDistill 各模块作用分析.}
我们在表~\ref{zh:tab_abliation_samad_all_modules} 展示了基于 SAM-AD  的 RoboDistill(L) 关键模块消融结果, 包括 AD-FPN、DGWA 与 KD Fusion. 实验表明, SAM-AD 能显著提升基线 Focals Conv~\upcite{zh:focalconv} 的性能, 从 (34.77\%, 41.30\%, 44.55\%, 80.97\%) 提升至 (80.68\%, 81.68\%, 81.67\%, 83.48\%). 进一步引入 AD-FPN、DGWA 与 KD Fusion 后, 模型性能持续提升, 验证了各模块的重要作用. 其中, AD-FPN 增强多尺度特征表达, DGWA 动态调节权重以优化模态融合, KD Fusion 实现跨模态知识迁移. 综上, 这些模块共同提升了 RoboDistill 在分布外噪声场景下的鲁棒性与泛化能力.

\subsection{可视化}
我们在 图 \ref{zh:fig:vis-kittic} 给出了KITTI-C 数据集的对比可视化结果,比较了我们的 RoboDistill(L) 与 LoGoNet 的表现.总体而言, 相比当前最先进的方法(例如 LoGoNet~\upcite{zh:logonet}), 我们的方法能够给出更准确的预测, 尤其是在对小目标和远距离目标的检测上表现更为突出. 通过利用视觉基础模型的泛化能力与鲁棒性, 我们的方法显著提升了多模态三维目标检测的鲁棒性, 并有效缓解了自动驾驶中的分布外(OOD)噪声场景带来的影响.

\section{结论}
本文提出了一种鲁棒且具备良好泛化能力的多模态3D目标检测框架---\textbf{RoboDistill}, 旨在应对自动驾驶场景中的分布外(OOD)噪声问题. 通过引入视觉基础模型(Visual Foundation Models, VFMs, 如 SAM)的强泛化能力, RoboDistill 有效缓解了传感器噪声、恶劣天气及环境变化带来的性能退化. 为适应自动驾驶任务, 我们提出了领域自适应预训练策略 \textbf{SAM-AD}, 并设计了 \textbf{AD-FPN} 模块对图像特征进行多尺度上采样与细化, 以实现与激光雷达特征的无缝融合. 同时, 引入的 \textbf{深度引导小波注意力模块(DGWA)} 有效抑制传感器噪声, 而 \textbf{多模态融合知识蒸馏(KD Fusion)} 模块通过将 VFMs 的高质量知识迁移至点云网络, 进一步增强了特征对齐与鲁棒性.
与前一版本 RoboFusion 相比, RoboDistill 在结构设计与知识蒸馏机制上均进行了显著改进, 并将模型的泛化能力扩展至 27 种分布外噪声场景. 基于 KITTI-C 与 nuScenes-C 基准的实验结果表明, RoboDistill 在多数复杂扰动下取得了更优或具有竞争力的性能, 展现出良好的鲁棒性与可扩展性. 总体而言, RoboDistill 有效弥合了干净基准性能与真实环境鲁棒性之间的差距, 为复杂场景下的多模态3D目标检测提供了可靠解决方案.

\noindent \textbf{局限性与未来工作.}
首先, RoboDistill 在很大程度上依赖于视觉基础模型的表征能力, 虽然这显著提升了基线模型的泛化性, 但也带来了较高的模型复杂度. 其次, 由于 SAM 与 FastSAM 的计算开销较大, RoboDistill(L) 与 RoboDistill(B) 的推理速度相对较慢; 相比之下, RoboDistill(T) 采用更轻量的 MobileSAM, 其推理速度更接近当前主流方法(如 TransFusion). 未来工作中, 我们将探索仅在训练阶段引入 SAM, 用以指导轻量级学生模型的学习, 从而进一步提升模型的实时性. 此外, 我们将进一步探索更多复杂且贴近真实世界的噪声场景, 以持续增强 RoboDistill 框架的鲁棒性与实用性.

\end{document}